\documentclass[journal]{IEEEtran}

\usepackage{microtype}
\usepackage{booktabs} % for professional tables
\usepackage{lineno,hyperref}

\usepackage{graphicx,subfigure}

\usepackage{bm}
\usepackage{url}
\usepackage{color}
\usepackage{mathrsfs}
\usepackage{amsmath}
\usepackage{amsfonts}
\usepackage{amssymb}
\usepackage{multirow}
\usepackage{caption2}
\usepackage{float}
\usepackage{hyperref}

\usepackage{algorithm}
\usepackage{algorithmic}

\def\1{\mathbf{1}}
\def\0{\mathbf{0}}

\def\X{{\bf X}}

\def\x{{\bf x}}

\def\v{{\bf v}}
\def\t{{\bf t}}

\def\y{{\bf y}}

\def\A{{\bf A}}
\def\B{{\bf B}}

\def\bbeta{\bm{\beta}}
\def\zzeta{\bm{\zeta}}
\def \SSigma{\bm{\Sigma}}
\def\u{{\bf u}}

\def\w{{\bf w}}

\def\I{{\bf I}}

\def\z{{\bf z}}

\newtheorem{assumption}{Assumption}
\newtheorem{remark}{ Remark}
\newtheorem{definition}{ Definition}
\newtheorem{lemma}{ Lemma}
\newtheorem{proposition}{ Proposition}
\newtheorem{theorem}{ Theorem}

\newcommand*\dif{\mathop{}\!\mathrm{d}}
\newenvironment{proof}{\hspace{0ex}\textsc{Proof}.\hspace{1ex}}{\hfill$\blacksquare$\newline}

\begin{document}

\title{Markov Subsampling based on Huber Criterion}

\author{Tieliang Gong,
        Yuxin Dong,
        Hong Chen, Bo Dong, Chen Li% <-this % stops a space
\thanks{T. Gong, Y. Dong, C. Li are with the School
of Computer Science and Technology, Xi'an Jiaotong University, Xi'an, Shaanxi 710049, China, (e-mail: adidasgtl@gmail.com; dongyuxin@stu.xjtu.edu.cn; cli@xjtu.edu.cn).}% <-this % stops a space
\thanks{B. Dong is with the School of Continuing Education,  Xi'an Jiaotong University, Xi'an, 710049, e-mail: dong.bo@mail.xjtu.edu.cn.}

\thanks{H. Chen is with the College of Science, Huazhong Agriculture University, Wuhan, 430070, email: chenh@mail.hzau.edu.cn.  }
}

\markboth{IEEE Transactions on Neural Networks and Learning Systems}%
{Shell \MakeLowercase{\textit{et al.}}: Bare Demo of IEEEtran.cls for IEEE Journals}

\maketitle

\begin{abstract}
Subsampling is an important technique to tackle the computational challenges brought by big data. Many subsampling procedures fall within the framework of  importance sampling, which assigns high sampling probabilities to the samples appearing to have big impacts. When the noise level is high, those sampling procedures tend to pick many outliers and thus often do not perform satisfactorily in practice.  To tackle this issue, we design a new Markov subsampling strategy based on Huber criterion (HMS) to construct an informative subset from the noisy full data; the constructed subset then  serves as a refined working data for efficient processing. HMS is built upon a Metropolis-Hasting procedure, where the inclusion probability of each sampling unit is determined using the Huber criterion to prevent over scoring the outliers. Under mild conditions, we show that the estimator based on the subsamples selected by HMS is statistically consistent with a sub-Gaussian deviation bound. The promising performance of HMS is demonstrated by  extensive studies on large scale simulations and real data examples. 
\end{abstract}


% \fi
\IEEEpeerreviewmaketitle

\section{Introduction}
\label{sec:intro}

\IEEEPARstart{R}{apid} advancement in modern science and technology introduces data with extraordinary size and complexity, which brings great challenges to conventional machine learning and statistical methods. In the literature, two fundamental approaches have emerged to tackle the challenges of big data: one is the divide-and-conquer strategy 
\cite{zhang2015divide}, which involves partitioning the data into manageable segments, implementing a particular algorithm on these data segments in parallel, and synthesizing a global output by aggregating the segmental outputs; the other approach is the subsampling strategy \cite{dhillon2013new}, which involves selecting a representative subset from the full data as a surrogate, and obtaining an output through further analyzation of the surrogate.  The divide-and-conquer strategy usually relies on high computational power with computing clusters and is particularly effective when a dataset is too big to fit in one computer. However, it still consumes considerable computational resources and the access of distributed computational platforms are restricted by high cost. As a computationally cheaper alternative, subsampling  gains its merit for the situation, when the computational resources are limited.

The key task of subsampling is to effectively identify important samples  in order to maintain the essential information of the  full data. This task is particularly challenging for big data, which often comes with poor quality (high noise level) due to the uncontrolled collecting process. In the literature, informative sampling strategies are commonly adopted, where important samples are given high probabilities to be selected. During the last two decades, extensive studies have been concluded on informative sampling e.g. statistical leverage score method \cite{drineas2012fast,drineas2011faster,ma2015statistical,rudi2018fast}, gradient method \cite{zhu2016grad} and influence function method \cite{ting2018optimal} etc. Leverage score subsampling assigns the sampling probabilities proportional to a distance measure within the covariates. It does not take into account the response and hence is sensitive to outliers. Both gradient-based subsampling and influence function based subsampling are using the response together with the covariates to design sampling patterns, in which the probabilities are computed proportional to the quadratic loss gradient or influence function. Although they do avoid the interference of outliers to some extent,  the estimators calculated upon the associated subsamples are highly dependent on a reliable pilot model, which may be difficult to obtain in highly noisy setup. 

Huber criterion \cite{huber1992robust} provides an effective way  to deal with this situation. It is a hybrid of square loss for relatively small errors and absolute loss for relative large ones and hence is robust to heavy-tailed errors and outliers. Recent studies have shown the great potential of Huber criterion for robust estimation and inference. For example, \cite{lambert2011robust} proposed to combine the Huber criterion and adaptive penalty as lasso and shows that the resulting estimator is more robust than adaptive LASSO in prediction and variable selection tasks. \cite{wang2020new} developed data-driven Huber-type methods for regression tasks and establishes sub-Gaussian type concentration bounds for the Huber-type estimator. In \cite{sun2020}, the adaptive Huber regression method was proposed, which significantly outperforms least squares both in terms of mean and standard deviation. Besides,  it admits exponential type concentration bounds when the error variables have  finite moments. \cite{fan2019adaptive}  investigated the non-asymptotic consistency of $\ell_1$ regularized robust M-estimator with Huber loss under Markov chain setting. \cite{CHEN2020571} additionally investigated collinearity and explored the grouping effect in Huber regression. \cite{Meyer2021CVPR} proposed an alternative probabilistic interpretation of minimizing the Huber loss, which is equivalent to minimizing an upper-bound on the Kullback-Leibler (KL) divergence in Laplacian settings.  \cite{wang2022huber} achieved robust forecasting based on Huber criterion for both non-Gaussian and non-stationary data.

In light of these advances, we aim to design a robust subsampling procedure by adopting the Huber criterion. To this end,  this paper proposes a Markov subsampling strategy based on Huber criterion (HMS) for linear regression. The procedure is as follows:  we first obtain a rough estimator $\bbeta_0$ based on a simple pilot selection, which determines the importance of each sample by calculating the Huber loss; we then perform subsampling from the full data $\mathcal{D}$ to generate a subset $\mathcal{D}_S$ through Metropolis-hasting (MH) type procedure, where the sampling probability is assigned according to the Huber loss. By doing so, samples with large Huber loss are unlikely to be selected and hence the noisy samples and outliers are ruled out with high probability. Moreover, MH sampling procedure and its variants require a proposal distribution to specify the sample importance, which is crucial to the success (e.g. fast convergence rate) of these algorithms, as improper selection of proposal distribution may result in misleading estimates. Different from MH-type algorithms, HMS determines the sample importance directly by Huber criterion, where the turning parameter is pre-specified through data-driven strategy, hence avoids such a problem. 

Our contributions are summarized as follows:
\begin{itemize}
    \item We develop a distribution-free Markov subsampling strategy based on Huber criterion to construct an informative subset from the noisy full data, which further enables robust statistical inference and prediction.
    \item Theoretically, we establish the statistical consistency for the regression estimator based on the subsample suggested by HMS in terms of Bahadur type representation \cite{bahadur1966note,he1996general}. Our results indicate that, with an appropriate robust parameter, the HMS-based estimator achieves nearly optimal convergence rate. The theoretical results also extends the error analysis of Huber estimator under i.i.d. samples to Markov dependent setup.
    \item Extensive empirical studies verify our theoretical findings. The promising performance of HMS estimator is also supported by both large-scale simulations and real data examples. 
\end{itemize}

The rest of the paper is organized as follows. Sections \ref{sec:pre} sets the notations  and problem statement. Section \ref{sec:huber} introduces the proposed Markov subsampling algorithm based on the Huber criterion. Section \ref{sec:theory} establishes the asymptotic analysis and the corresponding error bounds of the subsampling estimator. Section \ref{sec:exp} demonstrates experimental results on both simulation studies and real data examples. Section \ref{sec:con} concludes our work.

\section{Notations and Preliminaries} \label{sec:pre}
\subsection{Notations}
To make our arguments in the following section precise,  some concepts and notations  being used throughout this paper are introduced. 

Let $\u = (u_1, u_2, \cdots, u_d )^{\top} \in \mathbb{R}^d$  and $p \geq 1$, we denote the $\ell_p$-norm  and $\ell_{\infty}$-norm of $\u$ as $\|\u\|_p =( \sum_{i=1}^{d} |u_i|^p)^{1/p}$, $\|\u\|_{\infty} = \max_{j\in [1,d]} |u_j|$. For any $\w \in \mathbb{R}^d$, $\langle \u, \w \rangle = \u^{\top} \w$. For two scalars $a, b$, let $a \wedge b  = \min\{a, b\}$ and  $ a \vee b = \max\{a, b\}$. Given a matrix $\A \in \mathbb{R}^{m \times n}$, the corresponding spectral norm is defined by $\|\A\| = \max_{\u \in \mathbb{S}^{n-1}} \|\A \u\|_2$, where $\mathbb{S}^{n-1}$ is the unit sphere in $\mathbb{R}^{n}$. If $\A \in \mathbb{R}^{n \times n}$, we denote the minimum and maximum eigenvalue of $\A$  by $\lambda_{\min}(\A)$ and $\lambda_{\max}(\A)$. For a function $f : \mathbb{R}^d  \rightarrow \mathbb{R}$, we denote its gradient vector by $\nabla f \in \mathbb{R}^d$. 

\begin{definition}\cite{vershynin2018high}
	A random variable $X \in \mathbb{R}$ is said to be sub-Gaussian with variance proxy $\sigma^2$ if $\mathbb{E}[X] = 0 $ and its moment generating function satisfies
	\begin{equation}
		\mathbb{E}[\exp(sX)] \leq \exp \left(\frac{s^2 \sigma^2}{2}\right),~ \forall s\in \mathbb{R}.
	\end{equation}
\end{definition} 

The following concepts are important in our theoretical analysis. 
Let $\{X_i\}_{i\geq 1}$ be a Markov chain on a general space $\mathcal{X}$ with invariant probability distribution $\pi$.
Let $P(x, \dif y)$ be a Markov transition kernel on a general space $(\mathcal{X}, \mathcal{B}(\mathcal{X}))$ and $P^*$ be its adjoint. 
Denote $\mathcal{L}_2(\pi)$ by the Hilbert space consisting of square integrable functions with respect to $\pi$. For any function $h: \mathcal{X} \rightarrow \mathbb{R}$, we write $\pi(h) :=\int h(x) \pi(\dif x)$. Define the norm of $h \in \mathcal{L}_{2}(\pi)$ as $\|h\|_{\pi} = \sqrt{ \langle h, h \rangle}$.
Let $P^t(x, \dif y ), (t \in \mathbb{N})$  be the $t$-step Markov transition kernel corresponding to $P$, then for $i \in \mathbb{N}, x \in \mathcal{X}$ and a measurable set $S$, $P^t(x, S) = \mathrm{Pr}(X_{t+i} \in S| X_i = x)$. 
Following the above notations, we introduce the definitions of ergodicity and spectral gap for a Markov chain.

\begin{definition}
	Let $M(x)$ be a  non-negative function. For an initial probability measure $\rho(\cdot)$ on $\mathcal{B}(\mathcal{X})$, a Markov chain is uniformly ergodic if 
	\begin{equation}\label{eq_ergodic}
		\|P^{t}(\rho, \cdot) - \pi(\cdot)\|_{TV} \leq T(x) t^n
	\end{equation}
	for some $T(x) < \infty$ and $t < 1$, where $\|\cdot\|_{TV}$ denotes total variation norm.
\end{definition}
A Markov chain is geometrically ergodic if (\ref{eq_ergodic}) holds for some $t<1$, which eliminates the bounded assumption on $T(x)$. The dependence of a Markov chain can be characterized by the absolute spectral gap, defined as follows. 

\begin{definition}
	(Absolute spectral gap) A Markov operator $P$ has a $\mathcal{L}_2$ spectral gap $1 - \lambda$ if  
	\begin{equation}
		\lambda(P) := \sup \left\{ \|P h\|_\pi: \|h\|_\pi =1, \pi(h) = 0 \right\} < 1.
	\end{equation}
\end{definition}
The quantity $1 - \lambda$ measures the convergence speed of a Markov chain towards its stationary distribution $\pi$ \cite{rudolf2011explicit}. A smaller $\lambda$ usually implies faster convergence speed and less variable dependence.

\subsection{Huber Regression}
In this paper, we consider the data generated from the following linear regression model
\begin{equation}\label{eq_dataGen}
	y_i = \langle \x_i, \bbeta^* \rangle + \varepsilon_i, \quad  i = 1,2, \cdots,n
\end{equation}
where $y_i$ is the response, $\x_i \in \mathbb{R}^d$ is the covariate, $\varepsilon_i$ is the error and $\bbeta^* \in \mathbb{R}^d$ is the regression coefficient. It is well known that the ordinary least square estimator  $\bbeta_{ols}$ for (\ref{eq_dataGen}) has a suboptimal polynomial-type deviation bound, which makes it inappropriate for large scale estimation and inference. The key factor lies in the sensitivity of square loss to outliers \cite{catoni2012challenging}. To overcome this drawback, Huber loss  \cite{huber1992robust,sun2020} is proposed for achieving robust estimation. 
The Huber loss is defined by 
\begin{equation} 
	\ell_{\tau} (x) = \begin{cases}
		x^2/ 2, & \textrm{if} \quad  |x|\leq \tau , \\
		\tau |x| - \tau^2/2, & \textrm{if} \quad  |x| > \tau, 
	\end{cases} 
\end{equation}
where $\tau > 0$ is the robustification parameter that controls the bias and robustness. This function is quadratic with small values of $\tau$ while grows linearly for large values of $\tau$. The specification of $\tau$ is critical in practical applications. Some recent studies on  deviation bounds of Huber regression \cite{sun2020,wang2020new} suggest that $\tau$ should  be adaptive with  the dimension of input space, the moment condition of the noise distribution and the sample size to achieve robustness and unbiasedness estimate.  Specifically, Sun et al. \cite{sun2020} obtained near-optimal deviation bounds of Huber regression for both low and high dimensional cases. These observations will motivate us to derive optimal bounds for HMS estimation.

Define the empirical loss function $L_{\tau} (\bbeta) = \frac{1}{n} \sum_{i=1}^n \ell_{\tau} (y_i - \langle \x_i, \bbeta \rangle)$. The object of Huber regression is to find an optimizer of the following convex optimization problem
\begin{equation} \label{eq_huber_opt}
	\bbeta_{\tau}^* = \arg \min_{\bbeta \in \mathbb{R}^d} L_{\tau}(\bbeta),
\end{equation}
which can be easily solved via the iteratively reweighted least square method \cite{holland1977robust}. 
Denote the derivative of Huber loss $\ell_{\tau}(x)$ as $\varphi_{\tau}$, i.e 
\begin{equation}
	\varphi_{\tau} = \textrm{sign}(x) \min \{|x|, \tau \}, \quad x\in \mathbb{R}.
\end{equation}
In this paper, we focus on the setting that $n \gg d$. Denote $\X_S$ by the subsample matrix produced by HMS and $\bar{\x} = \SSigma^{-1/2}\x$. Suppose that $\SSigma = \mathbb{E}_\pi (\X_S \X_S^{\top})$ is positive definite, the regression errors $\varepsilon_i$ satisfy $\mathbb{E}(\varepsilon_i|\x_i) = 0$ and $v_{i, \delta} = \mathbb{E}(|\varepsilon_i|^{1 +\delta}|\x_i) < \infty$. With this setup, we write
\begin{equation*}
	v_{\delta} = \frac{1}{n}\sum_{i=1}^n v_{i, \delta} \quad \textrm{and} \quad u_{\delta} = \min \{v_{\delta}^{1/(1+\delta)}, \sqrt{v_1} \}, \quad \delta>0.
\end{equation*}

\section{Markov subsampling based on Huber criterion} \label{sec:huber}

As discussed before, the currently used informative measures (leverage score, gradient, influence function) in subsampling may not reflect the real contribution of each sample in highly-noisy settings, hence the resulting estimator can be misleading. To alleviate this issue, we develop a Markov subsampling strategy based on Huber criterion (HMS) to achieve robust estimation. The core idea is to select the samples with small errors based on Huber criterion by Markov chain  Monte Carlo (MCMC) method. Concretely, HMS consists of three steps: 1) pilot estimation; 2) Huber loss calculation; 3) Markov subsampling. 

\begin{figure}
	\centering
	\includegraphics[scale=0.65]{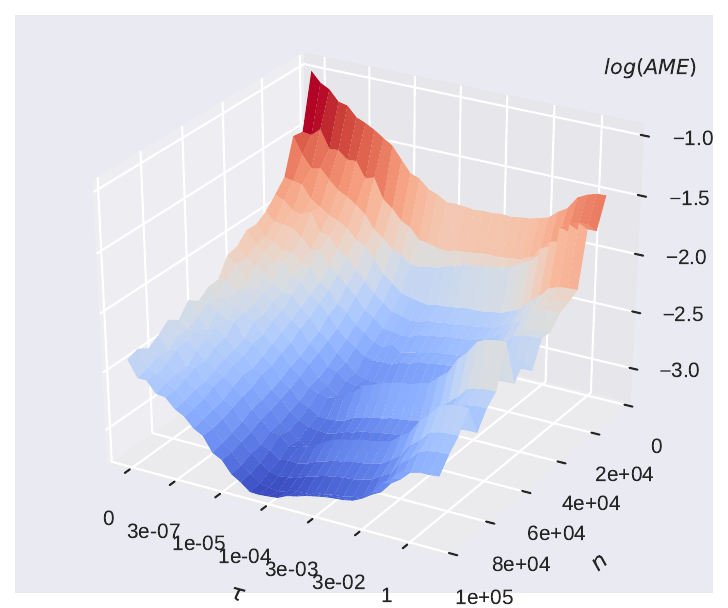}
	\caption{ $\log (AME)$  versus $\tau$ and $n$  (AME: Averaged Mean Error, defined in \ref{eq_AME}). Here, we generate  the data by (\ref{eq_dataGen}) with $n=1M$, $d= 500$ and $\varepsilon_i$ are i.i.d from student-t distribution with degree of freedoms  2.}
	\label{fig_phase}
\end{figure}

\begin{itemize}
	\item \textbf{Pilot estimation.}   The idea of pilot is widely applied in subsampling procedure  \cite{zhu2016grad,ting2018optimal,yu2020optimal,wang2019more}, where the sampling probability is specified by a pilot estimation. A popular way for calculating pilot is uniform subsampling.  To avoid bringing additional computational burden,  we suggest the pilot $\bbeta_0$ to be calculated by least square criterion based on a small random subset with user preference size $d< r \ll n$ , i.e.  $ \bbeta_0 = (\X_r^{\top} \X_r)^{-1} \X_r^{\top}\y_r$. It only takes additional $\mathcal{O}(rd^2)$ CPU time. We empirically demonstrate that HMS estimator does not  rely heavily on the quality of $\bbeta_0$.
	
	\item \textbf{Huber loss calculation.} The robustification parameter $\tau$ in Huber criterion plays a trade-off role between the bias and robustness. In practical, $\tau$ is usually set to be fixed through $95\%$ asymptotic efficiency rule \cite{huber1992robust,he1996general,rousseeuw2005robust,loh2017statistical}. However,  a fixed value may not  guarantee a good estimator, especially in highly noisy cases. As illustrated in Figure \ref{fig_phase}, $\tau$ should be adapted with $n,d$ (consider that  $n \gg d$, we ignore the effect of $d$).  It can be seen  that there  exists some $\tau$ such that the AME of $\bbeta_0$ achieves minimum for a fixed sample size $n$. In practical,  we first restrict $\tau$ in a reasonable range and select the optimal value then according to the minimal AME principal.  After specifying $\tau$, the importance of a sample $(\x_i, y_i)$ can be measured by the corresponding Huber loss $\ell_\tau(y_i - \x_i^{\top}\bbeta_0)$. The greater importance of a sample often comes with smaller Huber loss.
	
	\item \textbf{Markov subsampling.} It has been shown that the Markov chain samples may lead to  more robust estimation  than i.i.d counterparts in machine learning \cite{gong2015learning} and optimization tasks \cite{burke2020gradient,sun2018markov}. With this in mind we tend to implement probabilistic sampling through a Metropolis-Hasting type procedure. The core step,  probabilistic acceptance rule, is designed based on Huber criterion. Concretely, at some current sample $\z_t$, a randomly selected candidate sample $\z^*$ is accepted with probability defined in (\ref{eq_acc_prob}). If $\z^*$ is accepted, we set $\mathcal{D}_S = \mathcal{D}_S \cup \z^*$ and $\z^* = \z_{t+1}$. Otherwise, we randomly select a sample as a candidate and repeat this process. Finally, we accept the last $n_{sub}$ elements generated by this procedure after a user-specified burn-in period. 
\end{itemize}

The detailed procedures are summarized in Algorithm  \ref{alg_HMS}. Note that the
probabilistic acceptance rule (\ref{eq_acc_prob}) tends to select the samples with small Huber loss with high probability. Moreover,  the subsamples generated  by Algorithm \ref{alg_HMS} constitute an irreducible Markov chain, and therefore are uniformly ergodic \cite{meyn2012markov,down1995exponential}.  Computationally, HMS takes  $\mathcal{O}(n_0 d^2)$ time for pilot estimation, $\mathcal{O}((n_{sub} + t_0)d)$ time for Metropolis-Hasting sampling procedure and $\mathcal{O}(n_{sub}d^2 )$ time for optimizing (\ref{eq_huber_opt}) ( L-BFGS-B optimization strategy \cite{morales2011remark} is adopted). Hence, the total time complexity is $\mathcal{O}((2n_{sub} + t_0 )d^2 )$, which is much saving computational cost since $n_{sub} , t_0 \ll n$.

\begin{algorithm}[tb]
	\caption{Huber Regression with Markov Subsampling}\label{alg_HMS}
	\begin{algorithmic}[1]
		\STATE {\bfseries Input:}  Dataset $\mathcal{D} = (\x_i, y_i)_{i=1}^n$, subset $\mathcal{D}_S = \emptyset$, robustification parameter $\tau$, burn-in period: $t_0$, subsample size $n_{sub} \ll n$.

		\STATE Train a pilot estimator $\bbeta_0$ by $\bbeta_0 = (\X_r^{\top} \X_r)^{-1} \X_r^{\top}\y_r$, where $(\X_r, \y_r)$ are the random subsamples with size $n_0 = n_{sub}$. 
		
		\STATE Randomly select a sample $\z_1$ from $\mathcal{D}$, and set $ \mathcal{D}_s = {\z_1}$.
		
		\FOR{$2 \leq T \leq (n_{sub}+t_0)$}	
		\WHILE{$|\mathcal{D}_S | < T$}
		\STATE Randomly draw a candidate  $\z^* \!= \!(\x^*, y^*)$ 
		\STATE Calculate the acceptance probability by 
		\begin{equation} \label{eq_acc_prob}
			p = \min \left\{1,  \frac{ \ell_{\tau}( y_T - \langle \x_T, \bbeta_0 \rangle)}{\ell_{\tau} (y^* - \langle \x^*, \bbeta_0 \rangle)}\right \}
		\end{equation} 
		\STATE Set $
		\mathcal{D}_S = 
		\mathcal{D}_S \cup \z^* \ \text{with probability} \ p
		$
		\STATE If $\z^*$ is accepted, set $\z_{t+1} = \z^*$
		
		\ENDWHILE
		\ENDFOR
		
		\STATE Denote the last  $n_{sub}$ samples as $\mathcal{D}_S = (\X_S, \y_S) = \{ (\x_i, y_i)\}_{i=1}^{n_{sub}}$.
		
		\STATE Solve $\bbeta_\tau$ by Huber regression (\ref{eq_huber_opt}) based on $\mathcal{D}_S$.
		
		\STATE {\bfseries Output:}  $\bbeta_{\tau}$.
		
	\end{algorithmic}
\end{algorithm}

\begin{figure*}[h]
	{\footnotesize \ \centering
		\subfigure{
			\includegraphics[width = 7cm]{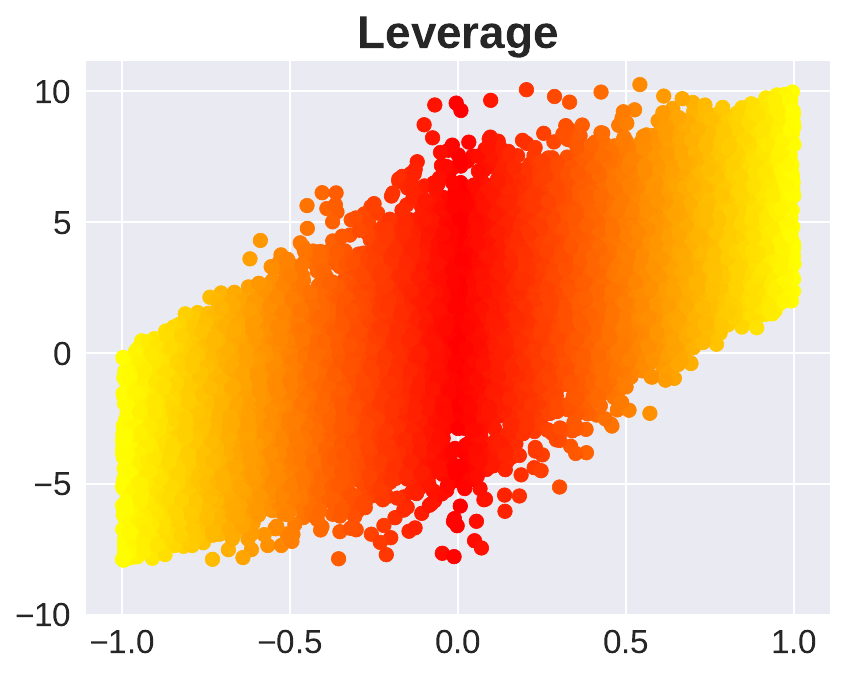}}
		\subfigure{
			\includegraphics[width = 7cm]{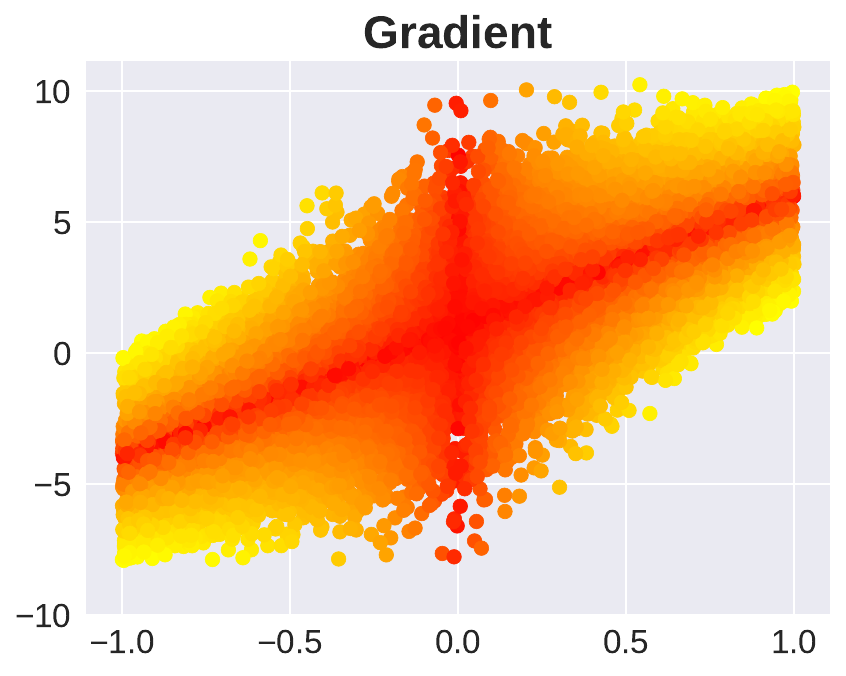}}
		\subfigure{
			\includegraphics[width = 7cm]{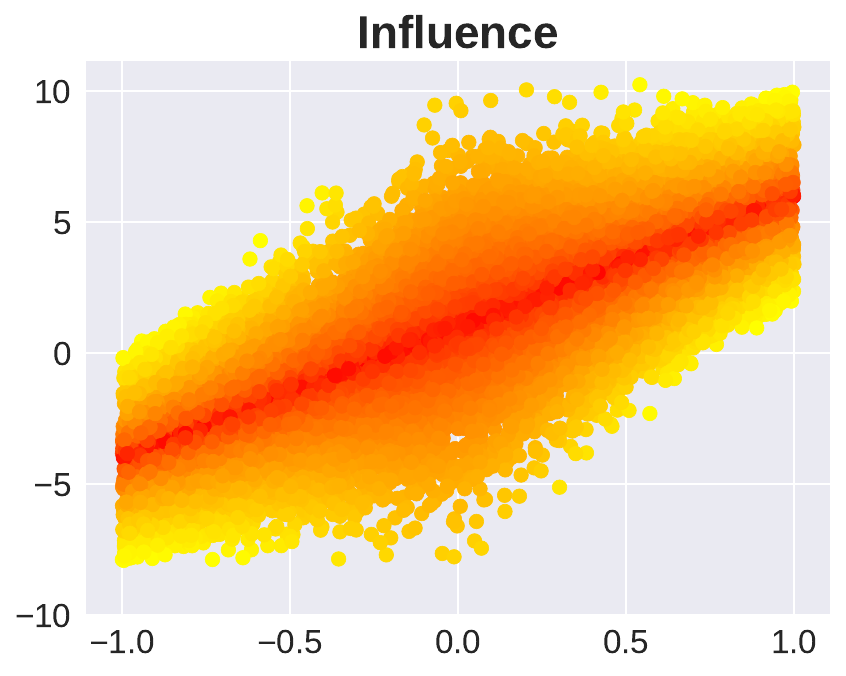}}
		\subfigure{
			\includegraphics[width = 7cm]{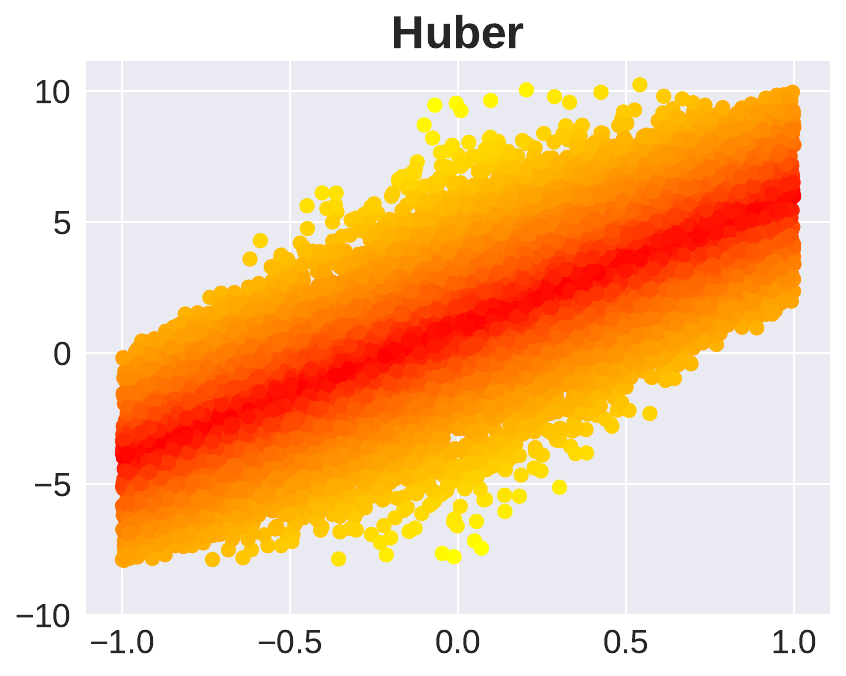}} 
		\caption{ \label{fig_prob}  \textit{I}llustration of sampling probabilities with different importance measures. The data are generated by $y = 5x + 1 + \varepsilon$ with $n = 10000$ and $\varepsilon$ is a mixture of Gaussian and uniform distribution. The bright yellow and red represent high and low sampling probability respectively.  Both leverage score and gradient are incline to select the points with large residuals near the center. The Influence tends to balance the regression design and the residual. The HMS can further enhance the effect of the influence.  }
	}
\end{figure*}

\section{Theoretical Assessments of HMS Estimator} \label{sec:theory}
In this section, we provide theoretical support for the proposed HMS. In particular, we aim at bounding the difference between the HMS estimator $\bbeta_{\tau}$ and the oracle $\bbeta^*$. Previous theoretical studies on subsample estimator are based on least squares \cite{dhillon2013new,ma2015statistical,zhu2016grad}, which has a closed-form solution. However, HMS estimator does not admit an explicit close-form representation and  the robustification parameter $\tau$ is not fixed,  all these pose the difficulties for analyzing its statistical properties.  To overcome these issues, we adopt the Lepski-type method developed in \cite{sun2020}.  We first present several necessary assumptions as below.

\begin{assumption} \cite{meyn2012markov} (Non-zero spectral gap Markov chain) \label{assumption1}
	The underlying Markov chain $\{X_i\}_{i=1}^n$ is stationary with unique invariant measure $\pi$ and admits a absolute spectral gap $1 - \lambda$.
\end{assumption}

\begin{assumption} (Bounded Covariates) \label{assumption2}
	There exists an envelop function $M: X \rightarrow \mathbb{R}$ such that for any function $f$, $\max |f(X)| \leq M(X)$ for $\pi$-almost every $X$.
\end{assumption}

\begin{assumption} (Bounded $(1+\delta)$-moments of errors) \label{assumption3}
	$\mathbb{E}(\varepsilon_i|X_i) = 0 $. For some $\delta > 0$ and $v_\delta >0$, $\mathbb{E}[|\varepsilon_i|^{1 + \delta}| X_i] < v_\delta$.
\end{assumption}

The absolute spectral gap $1- \lambda$ in Assumption \ref{assumption1} usually involves in spectral radius and geometrical ergodicity. Given the transition kernel $P$ of a Markov chain,  denote its spectral radius  by $\lambda_{\infty}(P) = \lim_{k\rightarrow \infty} \|P^k - \pi(\cdot)\|_{\pi}^{1/k}$. It is known that $\lambda_{\infty}(P) \leq \lambda(P)$ \cite{meyn2012markov}, where the equality holds for reversible Markov chain. The condition $1 - \lambda(P) >0$ implies geometrical ergodicity. A non-zero spectral gap is closely related to other convergence criterion of Markov chains \cite{roberts1997geometric}. Assumption \ref{assumption2} requires that the covariates are bounded by an envelop function, which can be a function of time, space or any forms of random variable. The boundedness assumption is quite common in statistics and learning theory analysis \cite{fan2018hoeffding,cucker2007learning}.  Assumption \ref{assumption3} requires errors to be with finite conditional $(1+\delta)$-moments, which covers a broad range of heavy-tailed noises including the student-t, the Pareto, log Normal and log Gamma et al. Now we are ready to present the main results for HMS estimator. 

The following Lemmas play an important role to prove our main theoretical results, where Lemma \ref{Lemma:Markov_Bernstein} is the Bernstein inequality within Markov-dependent setting, Lemma \ref{Lemma2}  gives the localized analysis on bounding $\beta_\eta$ and Lemma \ref{Lemma_convexity} presents the upper bound on the $\ell_2$ error between an estimation $\bbeta$ from a $d$-dimensional hypersphere and $\bbeta^*$. 

\begin{lemma} \cite{Jiang2018} \label{Lemma:Markov_Bernstein}
	Let $\{X_i\}_{i \geq 1}$ be a stationary Markov chain with invariant distribution $\pi$ and right $L_2$-spectral gap $1  - \lambda \in (0,1]$. Let $f_i : \mathcal{X} \rightarrow [-c, c]$ be a bounded function with $\pi(f_i) = 0$ and $\sigma^2 = \sum_{i=1}^n \pi(f_i^2)/n$. Then, for any $0 \leq t \leq (1 - \lambda ) /5c$, we have for any $\epsilon >0$,
	\begin{equation}
		\mathbb{P} \Big( \frac{1}{n} \sum_{i=1}^{n} f(X_i)   \geq \epsilon \Big) \leq \exp \Big( -\frac{n\epsilon^2}{2(A_1 \sigma^2 + A_2 c \epsilon)}  \Big), 
	\end{equation}
	where $A_1 = \frac{1 + \lambda}{ 1 - \lambda}$, $A_2 = \frac{\1}{3}\1_{\lambda =0} + \frac{5}{1  - \lambda} \1_{\lambda > 0} $. 
\end{lemma}

\begin{lemma} \label{Lemma2}
	\cite{fan2018lamm}
	Suppose $L$ is a convex  function. Let $D_L (\bbeta_1, \bbeta_2) = L(\bbeta_1) - L(\bbeta_2) - \langle  \nabla L(\bbeta_2), \bbeta_1 - \bbeta_2\rangle$ and $\bar{D}_L (\bbeta_1, \bbeta_2) = D_L(\bbeta_1, \bbeta_2) + D_L(\bbeta_2, \bbeta_1)$. For $\bbeta_{\eta} = \bbeta^* + \eta(\bbeta - \bbeta^*)$ with $\eta \in (0,1]$, 
	\begin{equation}
		\bar{D}_L(\bbeta_{\eta}, \bbeta^*) \leq \eta \bar{D}_L(\bbeta, \bbeta^*).
	\end{equation}
\end{lemma}

\begin{lemma} \cite{sun2020} \label{Lemma_convexity}
	Suppose $v_{\delta} < \infty$ for some $ 0 < \delta \leq 1$ and $(\mathbb{E}\langle \u, \bar{\x}  \rangle^4)^{1/4} \leq C \|\u\|_2$ for all $\u \in \mathbb{R}^d$ and some constant $C>0$. Moreover, let $\tau, r >0$ satisfy $\tau \geq  2\max\{ (4v_{\delta})^{1/ (1+\delta)}, 4 C^2 r \}$ and $n \geq (\tau/r)^2(d +t)$. Then with probability at least $1 - e^{-t}$,
	\begin{equation}
		\langle  \nabla L_{\tau}(\bbeta) - \nabla L_{\tau}(\bbeta^*), \bbeta - \bbeta^* \rangle \geq \frac{1}{4} \|\SSigma^{1/2} (\bbeta - \bbeta^*)\|_2^2
	\end{equation}
	uniformly over 
	\begin{equation*}
			\bbeta \in \B_0(r) = \{ \bbeta \in \mathbb{R}^d:  \| \SSigma^{1/2}(\bbeta - \bbeta^*)\|_2 \leq r \}.  
	\end{equation*}
\end{lemma}

 Proposition 1 provides a concentration inequality for $\|\SSigma^{-1/2} \nabla L_{\tau}(\bbeta^*) \|_2$, which is fundamental to our theoretical analysis.
\begin{proposition} \label{prop:grad}
		Suppose the Markov chain samples generated by Algorithm \ref{alg_HMS} are with invariant distribution $\pi$ and satisfy Assumptions \ref{assumption1} - \ref{assumption3}, then for any $0 < \delta \leq 1$, 
	\begin{align}
		\|\SSigma^{-1/2} \nabla L_{\tau}(\bbeta^*) \|_2  \leq   \frac{4\sqrt{\pi} C_0 A_2 (d + t)\tau}{n_{sub}}  \nonumber \\
		+ 4C_0 \sqrt{\frac{A_1 v_{\delta} \tau^{1 - \delta} (d+t)}{n_{sub}}}  + v_{\delta} \tau^{-\delta}
	\end{align}
	holds with confidence at least $1 - 2e^{-t}$, where $A_1 = \frac{1 + \lambda}{ 1 - \lambda}$, $A_2 = \frac{\1}{3}\1(\lambda \leq 0) + \frac{5}{1  - \lambda} \1({\lambda > 0}) $.
\end{proposition}
\begin{proof}
	To bound $\|\SSigma^{-1/2} \nabla L_{\tau}(\bbeta^*) \|_2$, we first define a  random vector 
	\begin{align}
		\zzeta^* &= \SSigma^{-1/2} \{\nabla L_{\tau}(\bbeta^*) - \nabla \mathbb{E} L_{\tau}(\bbeta^*) \}  \nonumber \\ 
		 & =  -\frac{1}{n_{sub}} \sum_{i=1}^{n_{sub}} \{\zeta_i \bar{\x}_i - \mathbb{E} (\zeta_i \bar{\x}_i ) \},
	\end{align}
	where $\zeta_i = \varphi_{\tau}(\varepsilon_i)$, $\bar{\x}_i = \SSigma^{-1/2}\x_i$ with $\SSigma = \mathbb{E}(\x \x^{\top})$ being  positive. Assume that there exists a $1/2$-Net $\mathcal{N}_{1/2}$ of the unit sphere $\mathbb{S}^{d-1}$ in $\mathbb{R}^d$  with $|\mathcal{N}_{1/2}| \leq 2^d$ such that $\|\zzeta^*\|_2 \leq 2 \max_{\u \in \mathcal{N}_{1/2}} | \langle u, \zzeta^* \rangle |$. Without loss of generality, we assume $\x_i$'s are centralized.  By Assumption 2, we know that  $\x_i$ are  sub-Gaussian vectors, i.e. 
	\begin{equation}
		\mathbb{P} (| \langle \u, \bar{\x} \rangle| \geq p) \leq \exp( - p^2\|u\|_2^2/ C_0^2) 
	\end{equation}
	for any $\u \in \mathbb{S}^{d-1}$ and $p \in \mathbb{R}$, where $C_0$ is a positive constant. We  then have
	\begin{equation}
		\mathbb{E}|\langle \u, \bar{\x}_i \rangle|^k \leq C_0^k k \Gamma(k/2), \quad k \geq 1, 
	\end{equation}
	it  immediately implies 
	\begin{equation}
		\begin{split}
			\sum_{i=1}^{n_{sub}} \mathbb{E}(\zeta_i \langle \u, \bar{\x}_i \ \rangle)^2 &\leq 2C_0^2 \tau^{1- \delta} \sum_{i=1}^{n_{sub}} v_{i,1}  = 2C_0^2 n v_{\delta} \tau^{1-\delta}, \\
			\sum_{i=1}^{n_{sub}} \mathbb{E}(\zeta_i \langle \u, \bar{\x}_i \ \rangle)^k &\leq \frac{k!}{2}(C_0 \tau/2)^{k-2} 2C_0^2 n_{sub} v_{\delta} \tau^{1 - \delta} 
		\end{split}
	\end{equation}
      for $k \geq 3$.  Furthermore, by Assumption 3, we have
	\begin{equation}
		\mathbb{E}[ \varphi_{\tau}(\varepsilon)] = - \mathbb{E}[(\varepsilon - \tau) \1(\varepsilon >\tau)] + \mathbb{E}[(-\varepsilon - \tau)\1(\varepsilon < -\tau)].
	\end{equation}
	Thus for any $k>2$,
	\begin{equation*}
		|\mathbb{E} \varphi_{\tau}(\varepsilon) | \leq \mathbb{E}[ (|\varepsilon| - \tau) \1(|\varepsilon| > \tau)] \leq \tau^{1-k} \mathbb{E}[|\varepsilon|^k].
	\end{equation*}
	It follows from Lemma \ref{Lemma:Markov_Bernstein} with $c =\sqrt{\pi} C_0 \tau$ and $\sigma^2 = 2C_0^2  v_{\delta} \tau^{1-\delta}$ that 
	\begin{equation}
		\mathbb{P}  \left\{  |\langle  \u, \zzeta^*\rangle| \leq \frac{2\sqrt{\pi} C_0 A_2 \omega \tau}{n_{sub}} \!+\! 2C_0 \sqrt{\frac{A_1 v_{\delta} \tau^{t-1} \omega}{n_{sub}}}    \right\} \!\geq\! 1  - 2e^{-\omega}
	\end{equation}
	for $ \forall  \omega >0$. By taking the union bound over $\u \in \mathcal{N}_{1/2}$, the following inequality 
	\begin{equation}\label{eq:zeta}
		\|\zzeta^*\|_2 \leq \frac{4\sqrt{\pi} C_0 A_2 \omega \tau}{n} + 4C_0 \sqrt{\frac{A_1 v_{\delta} \tau^{1 - \delta} \omega}{n}} 
	\end{equation}
	holds with confidence at least $1  - 2^{d+1}\cdot e^{-\omega}$.
	Then we consider the deterministic part $\|\SSigma^{-1/2} \nabla \mathbb{E} L_{\tau}(\bbeta^*) \|_2$, by direct calculation
	\begin{equation*}
		\|\SSigma^{-1/2} \nabla \mathbb{E} L_{\tau}(\bbeta^*) \|_2 \leq \sup_{\u \in \mathbb{S}^{d-1}}  \frac{1}{n_{sub}}\sum_{i=1}^{n_{sub}} \mathbb{E}|\zeta_i \langle \u, \bar{\x}_i \rangle| \leq v_{\delta} \tau^{-\delta}.
	\end{equation*}
	Let $\omega = d+t$, by combining above inequality and  (\ref{eq:zeta}), we obtain the stated result. 
\end{proof}

\begin{theorem}\label{th1}
	Suppose the Markov chain samples generated by Algorithm \ref{alg_HMS} are with invariant distribution $\pi$ and satisfy Assumptions \ref{assumption1} - \ref{assumption3}, then for any $t >0$, with confidence at least $1 - 2e^{-t}$, the HMS estimator $\bbeta_{\tau}$ with $\tau = \frac{1}{A_\lambda} (\frac{n_{sub}}{d+t})^{\max \{\frac{1}{1+\delta}, \frac{1}{2} \}}$ satisfies
	\begin{equation}
		\|\bbeta_{\tau} - \bbeta^*\|_2 \leq C_1  \lambda_{\max} (\SSigma^{1/2}) A_\lambda \Big(\frac{d+t}{n_{sub}} \Big)^{\min\{\frac{\delta}{1+\delta}, \frac{1}{2} \}}
	\end{equation}
	provided that $n_{sub} \geq C_2(d+t)$, where $C_1, C_2 >0$ are the constants independent of $n$ and $d$, $A_\lambda = \max \left\{ \sqrt{\frac{1+\lambda}{1 - \lambda}},  \frac{1}{3} \1_{\lambda=0} + \frac{5}{1 - \lambda} \1_{\lambda >0} \right\}$. 
\end{theorem}
\begin{proof}
	To begin with, recall that $\B_0(r) = \{ \bbeta \in \mathbb{R}^d:  \| \SSigma^{1/2}(\bbeta - \bbeta^*)\|_2 \leq r \}$ for some $r >0$.  Define $\bbeta_{\tau, \eta} := \bbeta^* + \eta(\bbeta_{\tau}  - \bbeta^*) \in \B_0(r)$, where $\eta \in (0,1 ]$.  Then we know from Lemma 2 that
	\begin{align}
		&\langle \nabla L_{\tau}(\bbeta_{\tau, \eta})  - \nabla L_{\tau}(\bbeta^*), \bbeta_{\tau, \eta} - \bbeta^* \rangle \nonumber \\
		& \leq \eta \langle \nabla L_{\tau}(\bbeta_{\tau}) - \nabla L_{\tau}(\bbeta^*), \bbeta_{\tau} - \bbeta^* \rangle.
	\end{align}
	It is easy to see $\nabla L_{\tau}(\bbeta_{\tau})  = 0 $ due to the KKT condition.  According to mean value theorem, 
	\begin{align}
		&\nabla L_{\tau}(\bbeta_{\tau, \eta})  - \nabla L_{\tau}(\bbeta^*)  \nonumber\\
		&= \int_{0}^{1} \nabla^2 L_{\tau}(t\bbeta_{\tau, \eta} + (1 -t)\bbeta^* ) \dif t (\bbeta_{\tau, \eta} - \bbeta^*)
	\end{align} 
	Assume there exist a constant $c > 0 $ such that 
	\begin{equation*}
		\min_{\bbeta \in \mathbb{R}^d: \|\bbeta - \bbeta^*\|_2 \leq r} \lambda_{\min}(\nabla^2 L_{\tau}(\bbeta)) \geq C_0,
	\end{equation*}
	hence $	C_0 \|\bbeta_{\tau, \eta} - \bbeta^*\|_2^2 \leq \|\nabla L_{\tau}(\bbeta^*)\|_2 \cdot \|\bbeta_{\tau, \eta} - \bbeta^*\|_2$, reducing the result yields
	\begin{equation}
		\|\bbeta_{\tau, \eta} - \bbeta^*\|_2 \leq C_0^{-1} \|\nabla L_{\tau}(\bbeta^*)\|_2. 
	\end{equation}
	Since $\bbeta_{\tau, \eta}\in \B_0(r)$, according to Lemma \ref{Lemma_convexity} with $r = \tau/ (4C_0^2)$, we get
	\begin{equation} \label{eq:th1}
		\langle  \nabla L_{\tau}(\bbeta_{\tau, \eta}) - \nabla L_{\tau}(\bbeta^*), \bbeta - \bbeta^* \rangle \geq \frac{1}{4} \| \SSigma^{1/2}(\bbeta_{\tau, \eta} - \bbeta^*)\|_2^2
	\end{equation}
	with confidence at least $1 - e^{-t}$. Then by Proposition 1, 
	\begin{equation} \label{eq:th2}
		\begin{split}
		\|\SSigma^{-1/2} \nabla L_{\tau}(\bbeta^*) \|_2  &\leq   \frac{4\sqrt{\pi} C_0 A_2 (d + t) \tau}{n_{sub}}  + v_{\delta} \tau^{-\delta}  \\ 
		& \quad + 4C_0 \sqrt{\frac{A_1 v_{\delta} \tau^{1 - \delta} (d+t)}{n_{sub}}} \\
		& : = r^* 
	\end{split}
	\end{equation}
	holds with confidence at least $1 - e^{-t}$. Combining (\ref{eq:th1}) and (\ref{eq:th2}), we know that with confidence at least $1 - 2e^{-t}$,
	\begin{equation}
		\|\bbeta_{\tau, \eta} - \bbeta^*  \|_2 \leq 4 r^*
	\end{equation}
	provided $n \geq C_1(d+t)$, where $C_1 > 0$ is a constant depending only on $C_0$. The constructed estimator $\bbeta_{\tau, \eta}$ lies in the interior of the ball with radius $r$. By the construction in the beginning of the proof, this enforce $\eta = 1$ and thus $\bbeta_{\tau} = \bbeta_{\tau, \eta}$. This completes the proof.
\end{proof}

\begin{remark}
	Theorem \ref{th1} indicates that the HMS estimator  $\bbeta_{\tau} $ is consistent under moderate conditions, i.e. $\| \bbeta_{\tau} - \bbeta^*\| \rightarrow 0 $ as $n_{sub} \rightarrow \infty$. The founding condition requires that the Markov chain generated by algorithm \ref{alg_HMS} has absolute spectral gap. HMS almost trivially meets this condition since the corresponding Markov chain is uniformly ergodic, and hence geometrically ergodic. Moreover, the error bound of HMS only requires finite moments of  error $\varepsilon_i$, which is weaker than sub-Gaussian error condition in linear regression models for subsampling  \cite{dhillon2013new,mcwilliams2014fast,zhu2016grad}. We find that $\tau$ should adapt with subsample size $n_{sub}$, the input dimension $d$, the moments of error term and the dependence of underlying Markov chain. In particular, with an appropriate choice of $\tau$, the convergence rate of HMS estimator is with $\mathcal{O}\Big((\frac{d}{n_{sub}})^{\min\{\frac{\delta}{1+\delta}, \frac{1}{2}\}}\Big)$ decay, which matches the near-optimal deviations in i.i.d. case \cite{sun2020}. Note that the Markov dependence impacts on $\tau$ in the way that the subsample size $n_{sub}$ is discounted by a factor $A_\lambda$. In other words, in order to achieve $\tau$-adaptation effect, the required subsample size  increases with $A_\lambda$ when transferring from i.i.d. sample setup to Markov dependence setup. Furthermore, a small value for  $\lambda$ implies a fast convergence rate of HMS estimator.
\end{remark}

\begin{theorem}\label{th2}
	Under the same conditions with Theorem \ref{th1},  for any $t >0$, the HMS estimator $\bbeta_{\tau}$ with $\tau = \sqrt{\frac{1-\lambda}{1+\lambda}}  \Big(\frac{n_{sub}}{( d  + t)\log d}\Big)^{\frac{1}{2(1+\delta)}}$ satisfies 
	\begin{align} \label{eq_th2}
		\mathbb{P} \Bigg\{  \left\|\SSigma^{1/2}(\bbeta_{\tau} - \bbeta^*)  - \frac{1}{n_{sub}} \sum_{i=1}^{n_{sub}} \varphi_{\tau}(\varepsilon_i) \bar{\x}_i  \right\|_2 \nonumber \\
		\geq C_3  \sqrt{\frac{1+\lambda}{1 - \lambda}} \sqrt{\frac{(d+t)\log d}{n_{sub}}} \Bigg\} \leq 3 e^{-t}
	\end{align}
	provided $n_{sub} \geq C_4(d+t)$, where $C_3, C_4$ are  the constants independent $n$ and $d$.
\end{theorem}
\begin{proof}
	Let $r_1 = 4 r^*$, we know from the proof of Theorem 1 that
	\begin{equation}\label{eq:th2_1}
		\mathbb{P} \left\{ \bbeta_{\tau} \in \B_0(r_1)  \right\} \geq 1 - 2 e^{-t}
	\end{equation}
	provided $n_{sub} \geq C_1(d+t)$. Define  random process  $\Phi(\bbeta) = L_{\tau}(\bbeta) - \mathbb{E}L_{\tau}(\bbeta)$ and 
	\begin{equation}
		\Psi(\bbeta) = \SSigma^{-1/2} \{\nabla L_{\tau}(\bbeta) - \nabla L_{\tau}(\bbeta^*) \}  - \SSigma^{1/2}(\bbeta - \bbeta^*).
	\end{equation}
	Our goal is to bound $\|\Psi(\bbeta_{\tau})\|_2 = \| \SSigma^{-1/2} (\bbeta_{\tau} - \bbeta^*) + \SSigma^{-1/2}\nabla L_{\tau}(\bbeta^*)\|_2 $, the key step lies in bounding the supremum of  empirical process $\{\Psi(\bbeta) : \bbeta \in \B_0(r) \}$. To  achieve this goal, we need to bound $\mathbb{E}\Psi(\bbeta)$ and $\Psi(\bbeta) - \mathbb{E}\Psi(\bbeta) $. 
	
	Denote $\hat{\bbeta}$ as the convex combination of $\bbeta$ and $\bbeta^*$. By mean value theorem, we see that 
	\begin{equation}
		\begin{split}
			\mathbb{E}\Psi(\bbeta) &= \SSigma^{-1/2} \{\nabla \mathbb{E}L_{\tau}(\bbeta) - \nabla \mathbb{E}L_{\tau}(\bbeta^*) \} -  \SSigma^{1/2}(\bbeta - \bbeta^*) \\
			&= \{\SSigma^{-1/2}\nabla^2 \mathbb{E} L_{\tau}(\hat{\bbeta}) \SSigma^{-1/2} - \I_d  \} \SSigma^{1/2} (\bbeta - \bbeta^*),
		\end{split}
	\end{equation}
	hence 
	\begin{equation}
		\sup_{\bbeta \in \B_0(r)} \|\mathbb{E}{\Psi(\bbeta) } \|_2 \leq r \times \sup_{\bbeta \in \B_0(r)} \| \SSigma^{-1/2}\nabla^2 \mathbb{E} L_{\tau}(\hat{\bbeta}) \SSigma^{-1/2} - \I_d \|.
	\end{equation}
	We known from Assumption \ref{assumption2} that $\|\x_i\|_{\infty} \leq M(\x)$, where $M(\x) : \x \rightarrow \mathbb{R}$ is a envelop function. Consider $\bbeta \in \B_0(r)$ and $\u \in \mathbb{S}^{d-1}$, we have
	\begin{equation}
		\begin{split}
			&| \u^{\top} \{\SSigma^{-1/2}\nabla^2 \mathbb{E} L_{\tau}(\hat{\bbeta}) \SSigma^{-1/2} - \I_d \}\u | \\ 
			&= \frac{1}{n_{sub}}\sum_{i=1}^{n_{sub}} \mathbb{E} \left\{ \1\{y_i - \langle  \x_i, \bbeta\rangle \geq \tau\} \langle  \u, \bar{\x}_i \rangle^2  \right\}  \\
			& \leq 	\frac{1}{n_{sub}}\sum_{i=1}^{n_{sub}} \mathbb{E} \Big\{ (\1\{|\varepsilon_i| \geq \tau/2\}   \\
			& \quad +\1\{\x_i^{\top}(\bbeta - \bbeta^*) > \tau/2\})  \langle  \u, \bar{\x}_i \rangle^2     \Big\}\\
			& \leq \frac{1}{n_{sub}}\sum_{i=1}^{n_{sub}} \mathbb{E} \left\{ (\1\{|\varepsilon_i| \geq \tau/2\} + \1\{\|\x_i\|_{\infty}> \tau/2r\})  \langle  \u, \bar{\x}_i \rangle^2     \right\} \\
			& \leq \frac{1}{n_{sub}}\sum_{i=1}^{n_{sub}} \mathbb{E} \left\{ (\1\{|\varepsilon_i| \geq \tau/2\} + \1\{\|M(\x)> \tau/2r\})  \langle  \u, \bar{\x}_i \rangle^2     \right\} \\
			& \leq  2^{1+\delta}\sigma^2 \tau^{-1-\delta} v_\delta + \sqrt{\frac{A_1 \log d}{n_{sub}}} + 4CA_1 \sigma^4 r^2,
		\end{split}
	\end{equation}
	which implies 
	\begin{equation} \label{eq:th2_2}
		\sup_{\bbeta \in \B_0(r)} \|\mathbb{E}{\Psi(\bbeta) } \|_2 \leq 2^{1+\delta}\sigma^2 \tau^{-1-\delta} v_\delta +  \sqrt{\frac{A_1 \log d}{n_{sub}}} + 4CA_1 \sigma^4 r^2.
	\end{equation}
Next, we focus on bounding $\Psi(\bbeta) - \mathbb{E}\Psi(\bbeta)$. To this end, we first rewrite 
	\begin{equation}
		\Psi(\bbeta) - \mathbb{E}\Psi(\bbeta)  = \SSigma^{-1/2} \{ \nabla \Phi(\bbeta) -  \nabla\Phi(\bbeta^*) \}.
	\end{equation}
	Set  
	\begin{equation}
		\Delta= \SSigma^{1/2}(\bbeta - \bbeta^*)
	\end{equation}
	 and define the empirical process 
	 \begin{equation}
	 	\bar{\Psi}(\Delta) := \Psi(\bbeta) - \mathbb{E} \Psi(\bbeta).
	 \end{equation}
 It is easy to check that $\bar{\Psi}(0)=0$ and $\mathbb{E}\bar{\Psi}(\Delta) = 0 $. For any $\u, \v \in \mathbb{S}^{d-1} $ and $m \in \mathbb{R}$, 
	\begin{equation*}
		\begin{split}
			&	\mathbb{E} \{ m\sqrt{n}\u^{\top} \nabla_{\Delta} \bar{\Psi}(\Delta) \v \} \\
			&\leq  \prod_{i=1}^{n_{sub}}  \Bigg\{   1+ \frac{m^2}{n_{sub}}\mathbb{E} \Big[ \Big(\langle \u, \bar{\x}_i\rangle^2 \langle \v, \bar{\x}_i \rangle^2 + \mathbb{E}|\langle \u, \bar{\x} \rangle^2 \langle \v, \bar{\x} \rangle|^2\Big)  \\
			& \quad \times e^{\frac{|m|}{\sqrt{n_{sub}}} \big(|\langle \u, \bar{\x}_i\rangle \langle \v, \bar{\x}_i \rangle| + \mathbb{E}|\langle \u, \bar{\x} \rangle^2 \langle \v, \bar{\x} \rangle| \big) } \Big] \Bigg\} \\
			& \leq \prod_{i=1}^{n_{sub}} \Bigg\{ 1 + e^{\frac{|m|}{\sqrt{n_{sub}}}} \frac{m^2}{n_{sub}} \mathbb{E}\big[e^{\frac{|m|}{\sqrt{n_{sub}}}} |\langle \u, \bar{\x}_i\rangle \langle \v, \bar{\x}_i \rangle| \big]  \\
			& \quad +  e^{\frac{|m|}{\sqrt{n_{sub}}}} \frac{m^2}{n_{sub}} \mathbb{E}
			\Big[ \langle \u, \bar{\x}_i\rangle^2 \langle \v, \bar{\x}_i \rangle^2 e^{\frac{|m|}{\sqrt{n_{sub}}} |\langle \u, \bar{\x}_i\rangle \langle \v, \bar{\x}_i \rangle| }  \Big]  \Bigg\}\\ 
		\end{split}
	\end{equation*}
	\begin{equation}
		\begin{split}
			& \leq \prod_{i=1}^n \Bigg\{ 1 + e^{\frac{|m|}{\sqrt{n_{sub}}}} \frac{m^2}{n_{sub}} \max_{\w \in \mathbb{S}^{d-1}} \mathbb{E} \Big[ e^{\frac{|m|}{\sqrt{n_{sub}}}} \langle \w, \bar{\x} \rangle^2 \Big] \\
			& \quad+  e^{\frac{|m|}{\sqrt{n_{sub}}}} \frac{m^2}{n_{sub}} \max_{\w \in \mathbb{S}^{d-1}} \mathbb{E} \Big[ \langle \w, \bar{\x} \rangle^4 e^{\frac{|m|}{\sqrt{n_{sub}}}} \langle \w, \bar{\x} \rangle^2 \Big] \Bigg\}  \\
			& \leq \exp \Bigg\{  m^2 e^{\frac{|m|}{\sqrt{n_{sub}}}} \Bigg( \max_{\w \in \mathbb{S}^{d-1}} \mathbb{E} \left[ e^{\frac{|m|}{\sqrt{n_{sub}}}} \langle \w, \bar{\x} \rangle^2 \right]  \\
			&\quad +  \max_{\w \in \mathbb{S}^{d-1}} \mathbb{E} \Big[ \langle \w, \bar{\x} \rangle^4 e^{\frac{|m|}{\sqrt{n_{sub}}}} \langle \w, \bar{\x} \rangle^2 \Big] \Bigg)  \Bigg\}.
		\end{split}
	\end{equation}
	Recall that each $\x_i$ is sub-Gaussian random variable, hence there exist constants $A_3$, $A_4$ depend only on $C_0$ such that for any $|m| \leq \sqrt{n_{sub}/A_3} $, 
	\begin{equation}
		\sup_{\u, \v \in \mathbb{S}^{d-1} }\mathbb{E} \{ m\sqrt{n_{sub}}\u^{\top} \nabla_{\Delta} \bar{\Psi}(\Delta) \v \} \leq \exp \{A_4 m^2/2 \}.
	\end{equation}
	By Theorem A.3 in \cite{spokoiny2013} , we see that 
	\begin{equation} \label{eq:th2_3}
		\mathbb{P} \left\{\sup_{\bbeta \in \B_0(r)} \|	\Psi(\bbeta) - \mathbb{E}\Psi(\bbeta) \|_2 \leq 6A_4 r \sqrt{8d+2t}   \right\} \geq 1 - e^{-t}
	\end{equation}
	when $n_{sub} \geq A_4(8d + 2t)$. Combing (\ref{eq:th2_2}) and (\ref{eq:th2_3}) together, we get 
	\begin{align}
		\sup_{\bbeta \in \B_0(r_1)} &\|  \SSigma^{-1/2} \{\nabla L_{\tau}(\bbeta) - \nabla L_{\tau}(\bbeta^*) \}  \!-\! \SSigma^{1/2}(\bbeta - \bbeta^*) \|_2  \nonumber\\
	& \leq 2^{1+\delta}\sigma^2 \tau^{-1-\delta} v_\delta +  \sqrt{\frac{A_1 \log d}{n_{sub}}} \nonumber\\
		 &+ 4CA_1 \sigma^4 r_1^2 \tau^{-2}  + 6A_4\sqrt{\frac{8d + 2t}{n_{sub}}} r_1.
	\end{align}
	with confidence at least $1 - e^{-t}$. This together with (\ref{eq:th2_1}) yield the final result. 
\end{proof}
\begin{remark}
	Theorem \ref{th2} provides a non-asymptotic Bahadur representation  \cite{he1996general} for HMS estimator $\bbeta_{\tau}$ when the error terms have finite (1+$\delta$)-th moments. It further implies that the approximation of $\bbeta_{\tau} - \bbeta^*$ has a sub-exponential tail. For the truncated random variable $\varphi_{\tau}(\varepsilon)$, we can see that
	\begin{align}
		|\mathbb{E}\varphi_{\tau}(\varepsilon)| &= - \mathbb{E}[(\varepsilon - \tau) \1(\varepsilon >\tau)] + \mathbb{E}[(-\varepsilon - \tau)\1(\varepsilon < -\tau) ] \nonumber \\
		&\leq  \mathbb{E}[(|\varepsilon| - \tau) \1(|\varepsilon|>\tau)] \nonumber \\
		& \leq \tau^{1-\delta} \mathbb{E}(|\varepsilon|^{\delta}).
	\end{align}
	This together with (\ref{eq_th2})  show that 
	the HMS estimator $\bbeta_{\tau}$ achieves non-asymptotic robustness against to heavy-tailed noise. Specifically, by taking $$t  = \log (n_{sub}), \tau  \asymp \sqrt{\frac{1-\lambda}{1+\lambda}}\sqrt{ \frac{n_{sub}}{d+ \log(n_{sub})}  },$$ we have
	\begin{align}
		\left \|\bbeta_{\tau} - \bbeta^* - \frac{1}{n_{sub}} \sum_{i=1}^{n_{sub}} \varphi_{\tau}(\varepsilon_i) \SSigma^{-1}\x_i  \right\|_2 \nonumber \nonumber\\
		= \mathcal{O} \left(\sqrt{\frac{1+\lambda}{1 - \lambda}} \sqrt{\frac{d+ \log(n_{sub})}{n_{sub}}} \right).
	\end{align}
	with confidence at least $1 - \mathcal{O}(n_{sub}^{-1})$. From an asymptotic viewpoint, it  implies that if  $d = o(n_{sub})$ as $ n_{sub} \rightarrow \infty$, then for any deterministic vector $\u \in \mathbb{R}^d$, $\langle \u, \bbeta_{\tau} - \bbeta^* \rangle$ converges to $n_{sub}^{-1}\sum_{i=1}^{n_{sub}} \varphi_{\tau}(\varepsilon_i) \SSigma^{-1}\x_i $ in distribution.
\end{remark}

\begin{table}[h]
	\caption{Statistics of real-world datasets}
	\label{datasets}
	\begin{center}
		\par
			\begin{tabular}{c||cc}
				\hline
				Datasets &  \# Sample size   & \# Features \\ \hline
				Appliances Energy Prediction & 19735  & 29 \\
				Poker Hand & 25010 & 11 \\
				Gas Turbine CO and NOx Emission & 36733  & 11 \\
				Wave Energy Converters & 288000 &32 \\
				PPPTS  & 45730  & 9 \\
				Beijing Multi-Site Air-Quality& 382168  & 14 \\
				\hline
		\end{tabular}
	\end{center}
\end{table}

\section{Experimental Results}\label{sec:exp}
This section aims to evaluate the empirical performance of the proposed HMS procedure. All numerical studies are implemented with Python 3.8 under Ubuntu 16.04 operation system with 2.2 GHz CPUs and 256 GB memory.

\begin{figure*}[h]
	{\footnotesize \ \centering
		\subfigure{
			\includegraphics[scale=0.4]{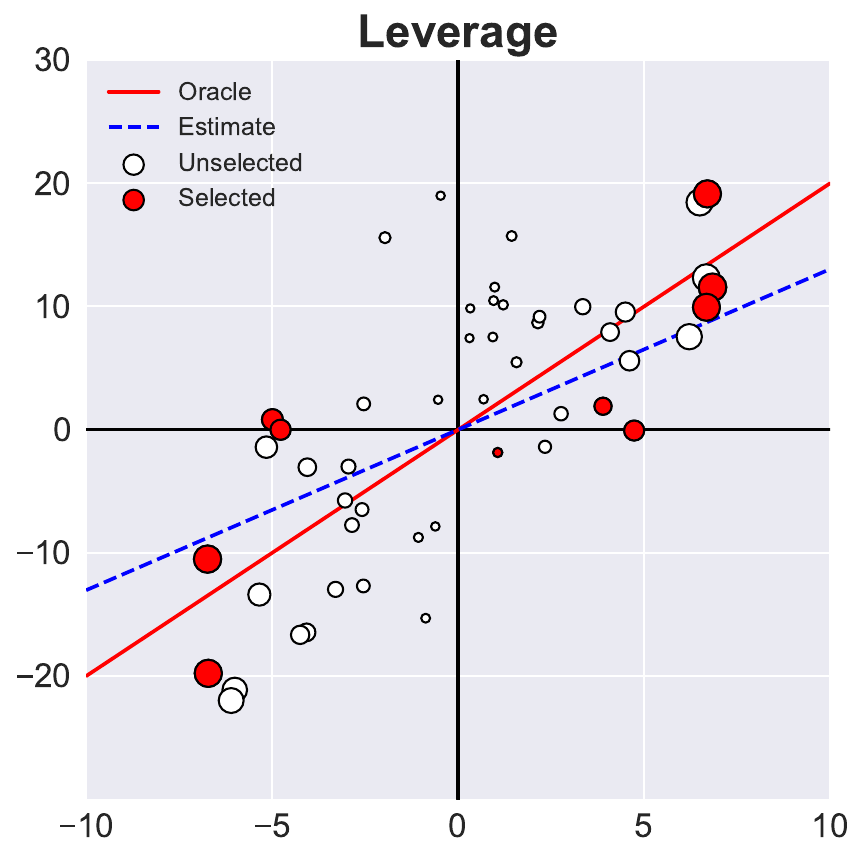}}
		\subfigure{
			\includegraphics[scale=0.4]{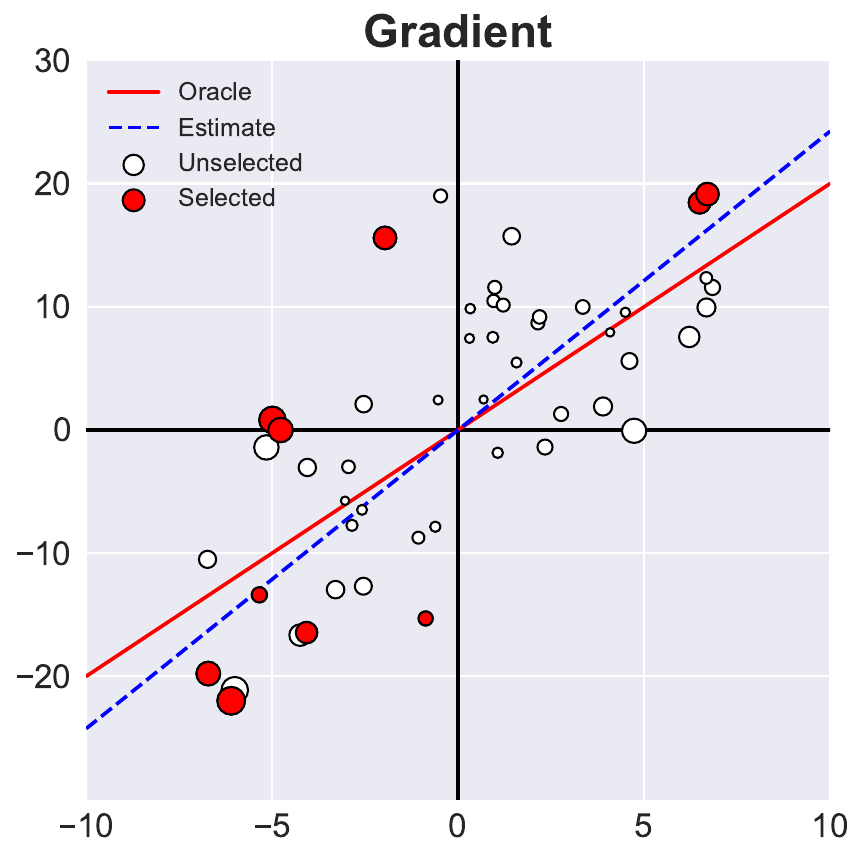}}\\
		\subfigure{
			\includegraphics[scale=0.4]{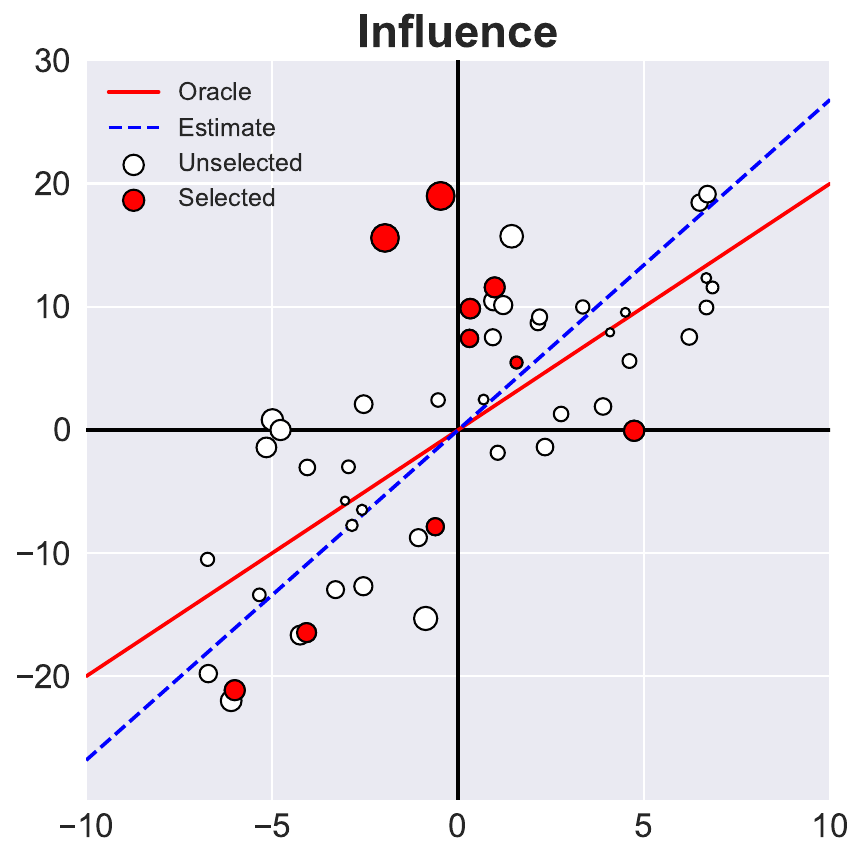}}
		\subfigure{
			\includegraphics[scale=0.4]{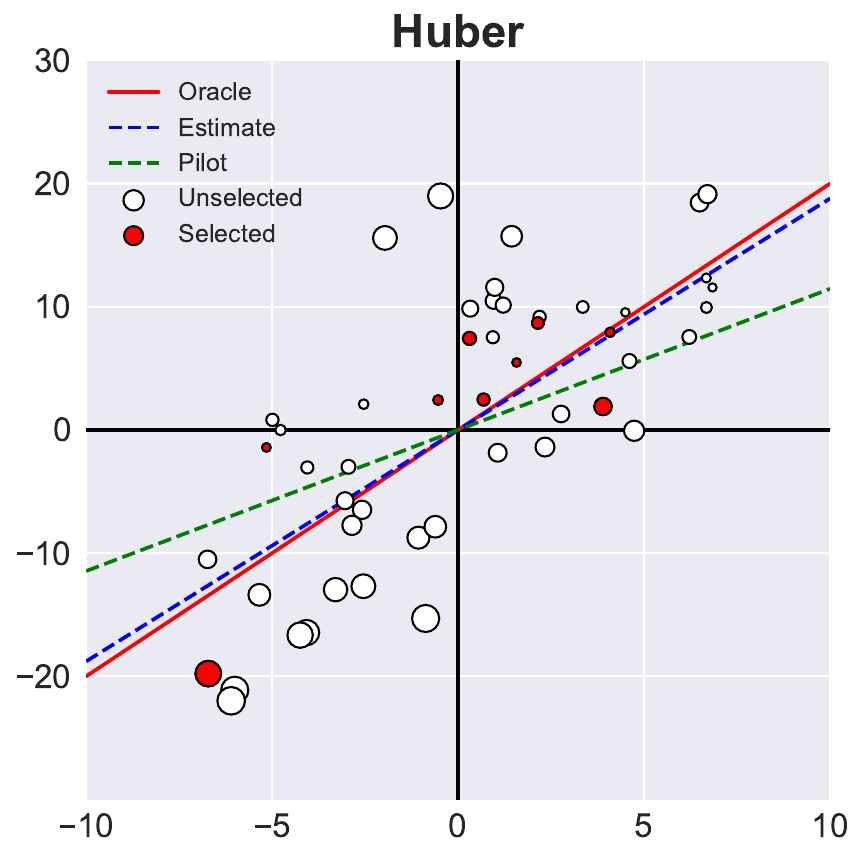}} 
		\caption{ \label{fig_pilot}  Comparisons on different sampling patterns. The oracle, pilot and subsampled estimator are denoted by the red real line, the green dashed line and the blue dashed line respectively. }
	}
\end{figure*}

\subsection{Sampling Pattern}
We first investigate the performance of HMS through comparing the sampling pattern  to leverage sampling, gradient-based sampling (GS) and influence-based sampling (IS).  A toy data is generated by $y = 2 x+ \varepsilon$ with $n=50, d=1$, where noise term comes from the student's $\t$ distribution with $2$ degrees of freedom, i.e. $\varepsilon \sim \t(2)$. Considering that both GS and IS require a pilot to determine the sampling probability, here we fix the pilot (marked by green dashed line) for a fair comparison.  The pilot is specified by uniform sampling $n_0 = 10$ points.  The turning parameter $\tau$ of HMS is set to $0.1$.   We plot $n_{sub} =10$ data points  (marked in red) selected by different sampling approaches, where the size denotes the corresponding assigned sampling probability. The estimators of four sampling approaches are then calculated  based on the subsampled data. As illustrated in Figure \ref{fig_pilot}, we see that the selected data points of HMS are more close to the oracle (marked by red real line) than competitors, hence the subsampled estimator (marked by blue dashed line ) can better recover the ground-truth estimator. Moreover, it can be observed that HMS can return a reliable estimator even the pilot is deviated from the oracle, which implies its great potential on selecting informative data from the noisy data. 

\subsection{Phase Transition}
\begin{figure*}[h]
	\centering
	\includegraphics[scale=0.5]{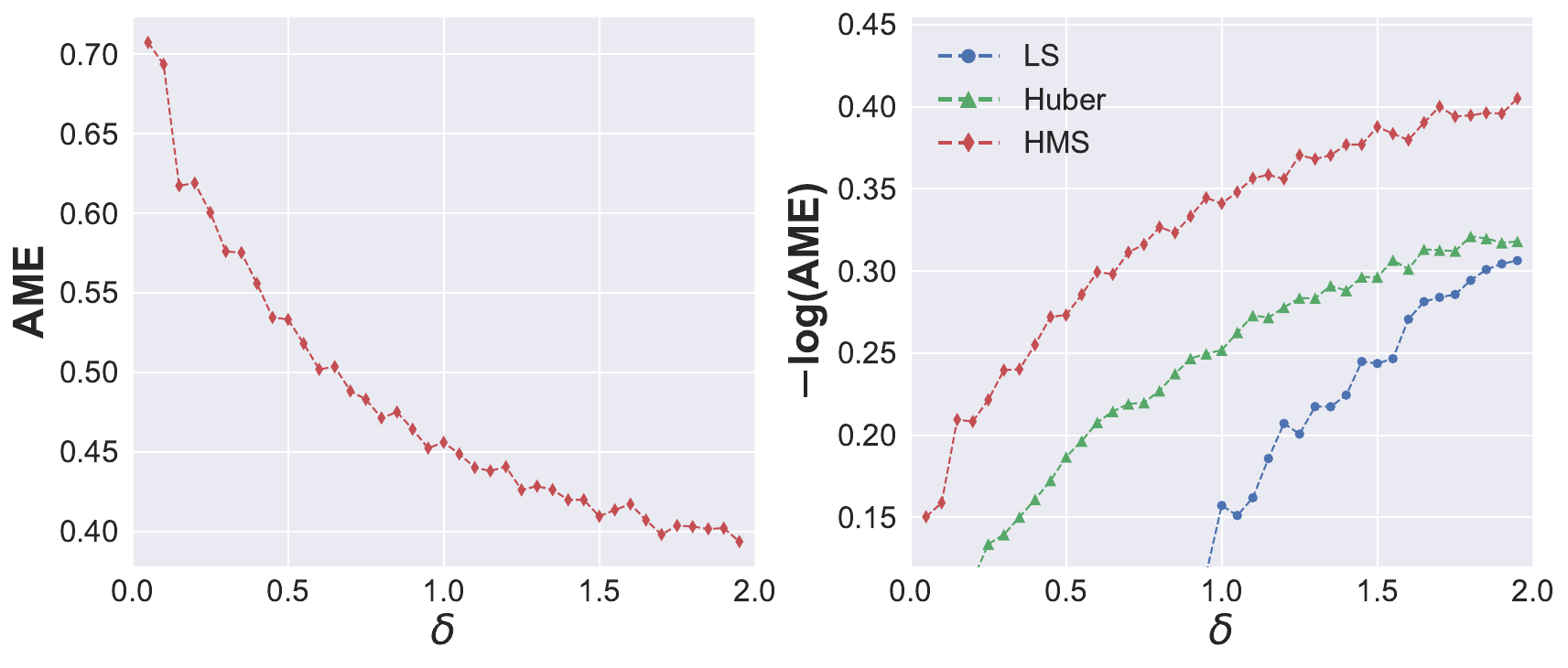}
	\caption{Left: AME of HMS. Right: Comparisons on $-\log(\textrm{AME})$ of different sampling procedures.}
	\label{fig_phase_tran}
\end{figure*}

Theorem (\ref{th1}) implies that 
\begin{align*}
	-\log (\|\bbeta_\tau - \bbeta^*\|) \asymp \frac{\delta}{1 + \delta} \log(n_{sub}) - \frac{\delta}{1+\delta} \log(A_\lambda v_\delta),  \\   0 < \delta \leq 1.
\end{align*}
In order to validate the phase transition behavior of HMS estimator, we  generate the data by (\ref{eq_dataGen}) with $n = 10K$, $d= 50$ and sample independent noise from $\t(df)$ , which has finite $(1 + \delta)$-th moments provided $\delta < df - 1$ and infinite $df$-th moment. The oracle $\beta^*$ is generated from discrete uniform distribution $\{\pm 3, \pm 2, \pm 1, 0\}$.  Following the setting in \cite{sun2020}, we set $n_{sub} = 1000$, $\delta = df - 1- 0.05$. The turning parameter $\tau$ is specified by $\tau = \sigma \sqrt{n_{sub}/ t}$, where $\sigma^2 = \frac{1}{n}\sum_{i=1}^{n}(y_i - \bar{y})$ with $\bar{y} = \frac{1}{n}\sum_{i=1}^n y_i$. The quality of the fit is measured by the absolute mean error (AME): 
\begin{equation} \label{eq_AME}
	\textrm{AME} = \frac{1}{K} \sum_{k=1}^{K} \|\bbeta_{\tau k} - \bbeta^*\|.
\end{equation}
Figure \ref{fig_phase_tran} displays the AME comparisons for HMS, least square with uniform sampling and Huber regression with uniform sampling. One can observe that the AME of HMS estimator is decreasing with the increase of $\delta$. In particular,  HMS can achieve lower AME than  Huber and LS with the varying degrees of freedom.  This further exhibits the significant advantages of HMS in robust regression. 

\begin{figure*}[h]
	\centering
	\includegraphics[scale=0.5]{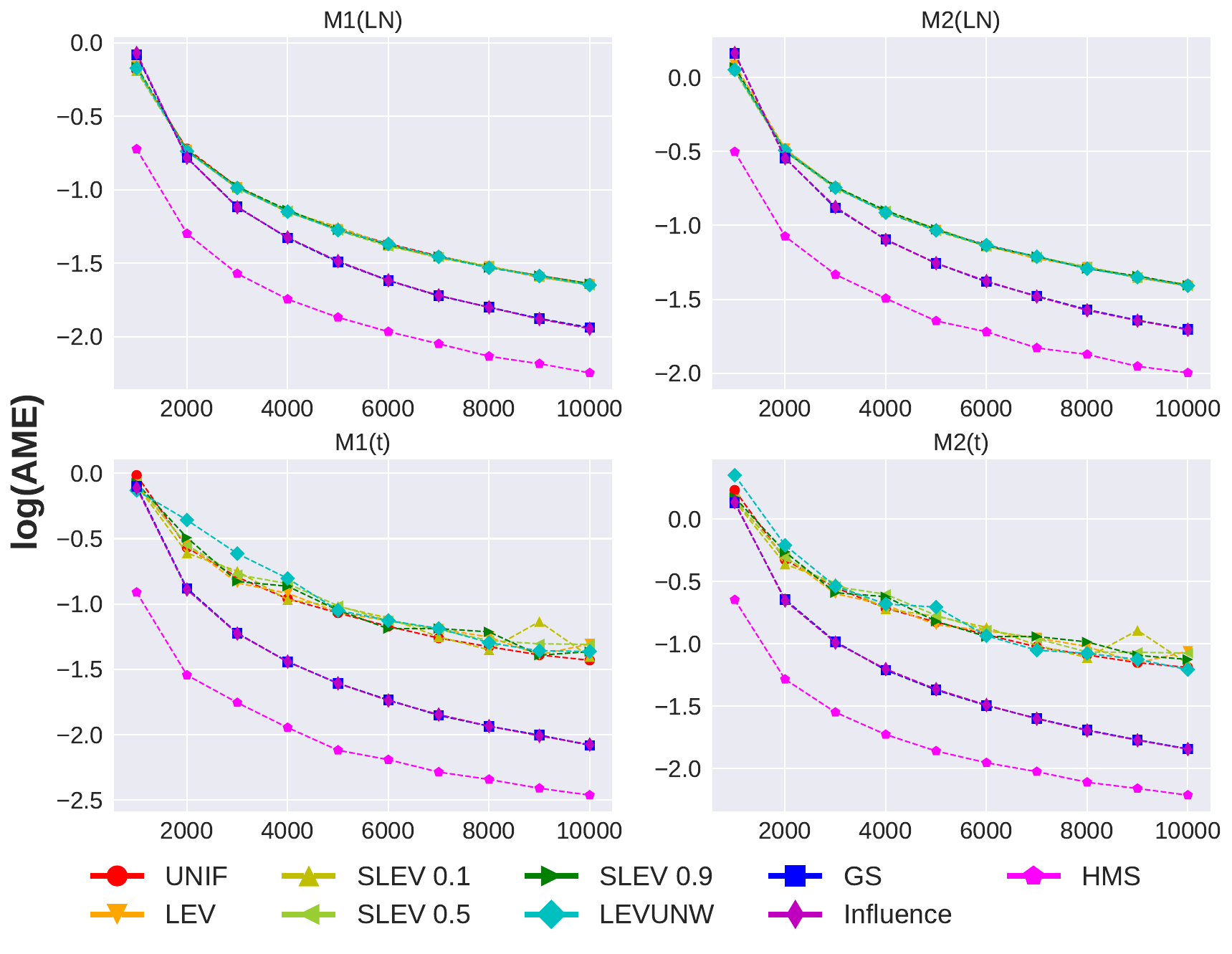}
	\caption{Comparisons on AME of different sampling procedures. In all settings, we vary the subsample size $n_{sub} = sr *n$ with $sr = [0.002, 0.004, 0.006,0.008, 0.01]$. }
	\label{fig_line}
\end{figure*}

\begin{table*}[th]
    \centering
    \setlength{\tabcolsep}{4pt}
    \begin{tabular}{c||ccc||ccc||ccc}
        \hline
        \multirow{2}{*}{Methods} & \multicolumn{3}{c||}{$n = 100K, d = 50$} & \multicolumn{3}{c||}{$n = 500K, d = 250$} & \multicolumn{3}{c}{$n = 1M, d = 500$} \\
        \cline{2-10}
        & $sr = 0.1\%$ & $sr = 0.5\%$ & $sr = 1\%$ & $sr = 0.1\%$ & $sr = 0.5\%$ & $sr = 1\%$ & $sr = 0.1\%$ & $sr = 0.5\%$ & $sr = 1\%$ \\
        \hline
        LEV & 55.9 & 56.1 & 57.6 & 1515.1 & 1597.5 & 1624.7 & 12345.4 & 12068.4 & 12947.8 \\
        SLEV & 54.1 & 56.8 & 62.9 & 1543.8 & 1610.9 & 1640.5 & 9932.8 & 9106.2 & 9545.5 \\
        LEVUNW & 53.8 & 55.6 & 61.1 & 1550.0 & 1580.1 & 1625.4 & 8235.8 & 8984.8 & 8918.5 \\
        GS & 33.7 & 34.0 & 37.6 & 818.2 & 804.2 & 879.2 & 3505.3 & 3473.6 & 3526.8 \\
        IS & 64.8 & 63.5 & 79.1 & 1618.5 & 1645.9 & 1794.8 & 8852.1 & 9243.0 & 10051.1 \\
        HMS & 5.9 + 23.2 & 6.6 + 34.5 & 7.5 + 44.4 & 141.1 + 229.2 & 167.1 + 302.5 & 189.0 + 389.2 & 557.9 + 780.8 & 662.4 + 910.0 & 754.9 + 1076.4 \\
        \hline
    \end{tabular}
    \caption{Comparisons on time cost (milliseconds) for different sampling methods. The time cost of HMS consists of two parts: selection of $\tau$ (left) + sampling (right).}
    \label{tab_time}
\end{table*}

\begin{figure*}[h]
	\centering
	\includegraphics[scale=0.45]{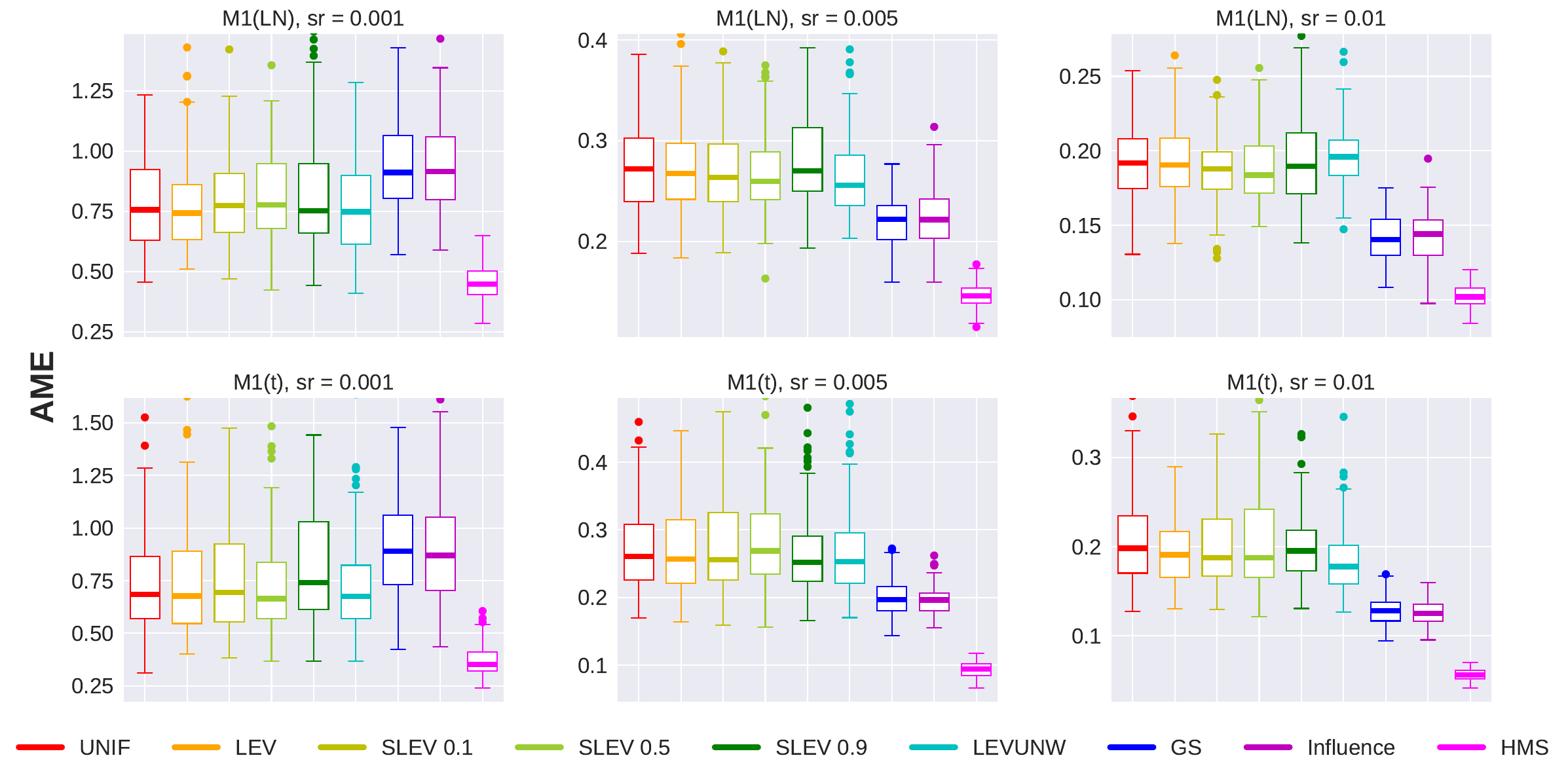}
	\caption{Boxplots of AME for different subsampling methods ($n = 10K, d=50$) }
	\label{fig_boxplot1}
\end{figure*}

\begin{figure*}[h]
	\centering
	\includegraphics[scale=0.45]{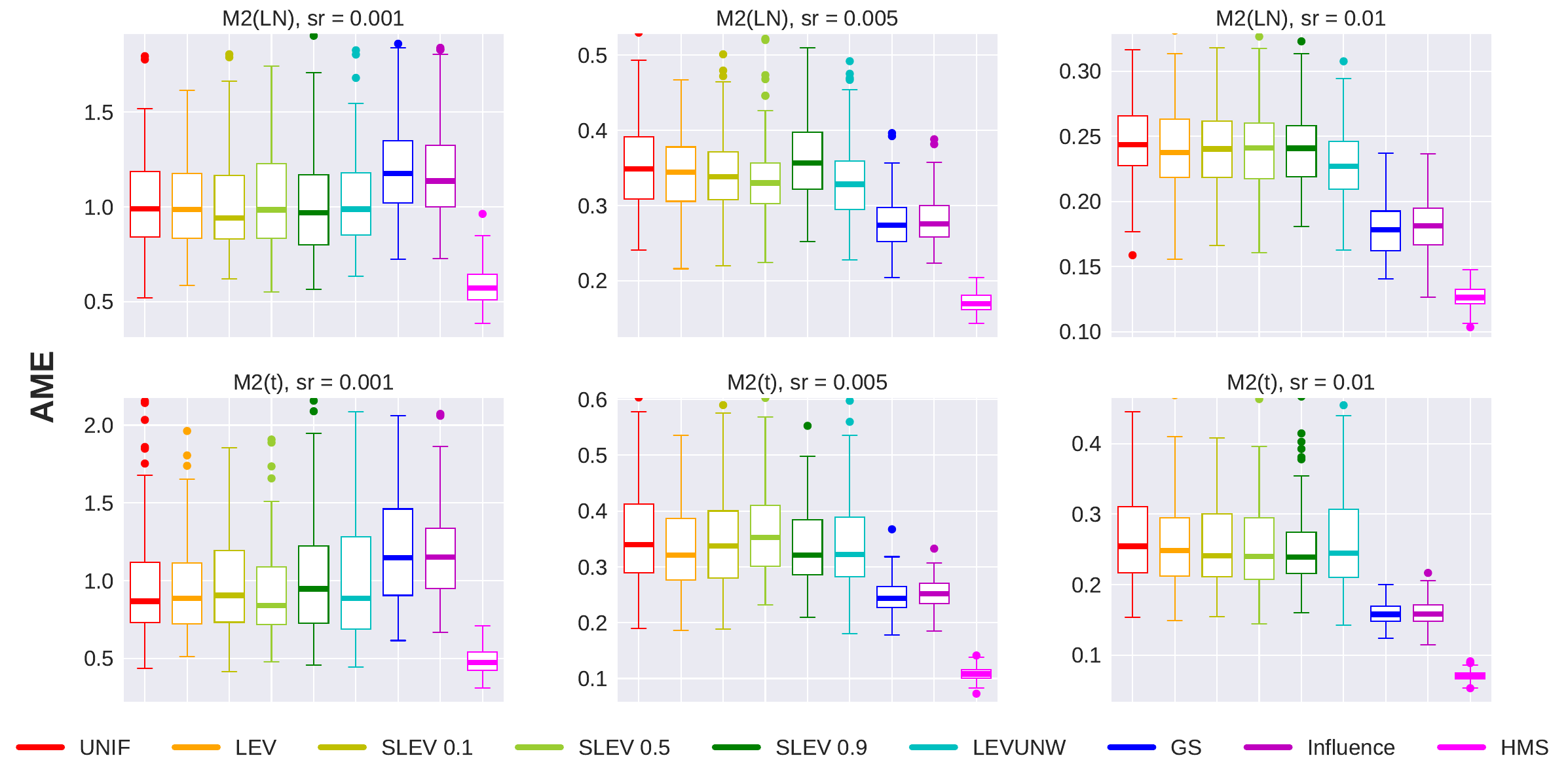}
	\caption{Boxplots of AME for different subsampling methods ($n = 10K, d=50$)}
	\label{fig_boxplot2}
\end{figure*}

\begin{figure*}[h]
	\centering
	\includegraphics[scale=0.45]{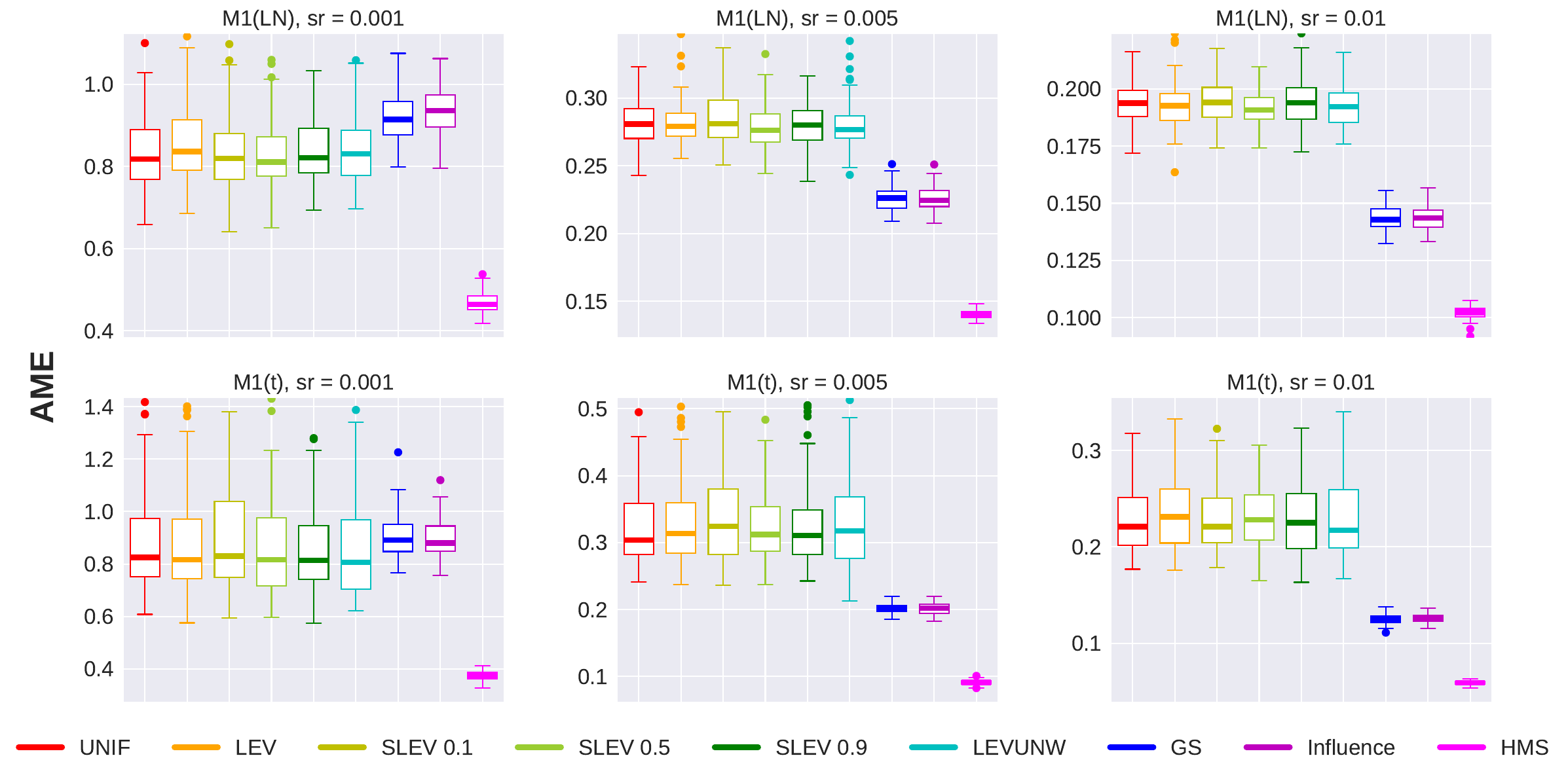}
	\caption{Boxplots of AME for different subsampling methods  ($n = 1M, d=500$)}
	\label{fig_boxplot3}
\end{figure*}

\begin{figure*}[h]
	\centering
	\includegraphics[scale=0.45]{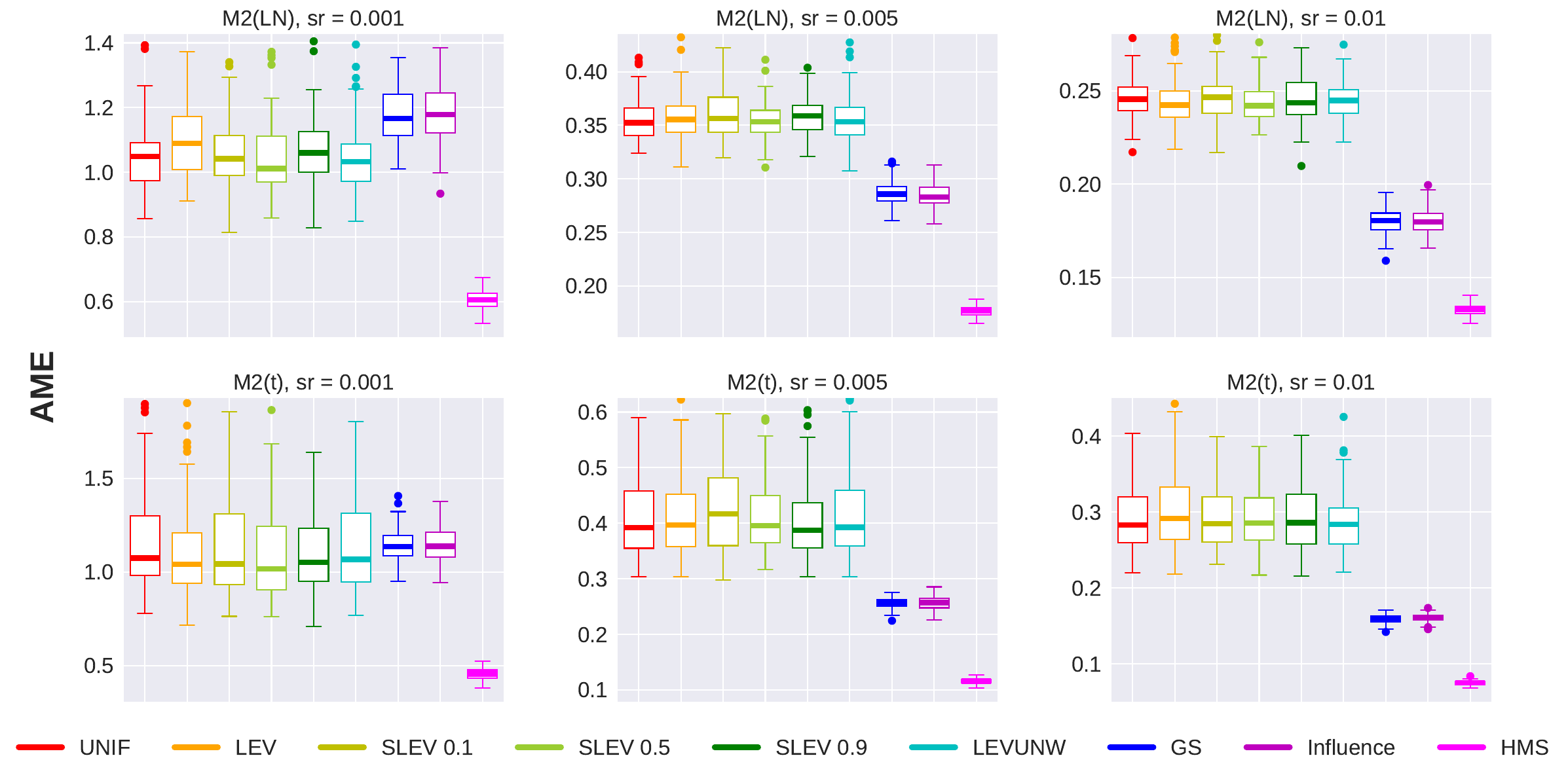}
	\caption{Boxplots of AME for different subsampling methods ($n = 1M, d=500$)}
	\label{fig_boxplot4}
\end{figure*}

\subsection{Simulation Studies}

We generate the data by $ \y  = \X \bbeta^*  + \bm{\varepsilon}$ \cite{zhu2016grad},
where the $n \times d$ design matrix $\X$ is constructed by a mixture of Gaussian  $\frac{1}{2} \bm{N}(\mu_1, \sigma_1^2) + \frac{1}{2} \bm{N}(\mu_2, \sigma_2^2)$ in two different ways: $(M1) \mu_1 = -2, \sigma_1 = 3, \mu_2 = 2, \sigma_2 = 10$; $(M2) \mu_1 = 0, \sigma_1 = 3, \mu_2 = 0, \sigma_2 = 10$. We generate two different types of i.i.d noise, including log-normal distribution $\varepsilon_i \sim \textrm{Lognormal}(0, 1)$ and Student-t distribution $\varepsilon_i \sim \bm{t}(2)$, both of them are heavy tailed and produce outliers with large variance.  We denote the models combining these design matrices and noise distributions as follows: $M1(\bm{LN}), M1(\bm{t}), M2(\bm{LN}), M2(\bm{t})$. 

\begin{table*}[h]	
	\begin{center}
		\resizebox{\textwidth}{14mm}{
			\footnotesize
			\begin{tabular}{ c||ccc||ccc||ccc }
				\hline
				\multirow{2}{4em}{Methods} & \multicolumn{3}{|c||}{Appliances Energy Prediction} & \multicolumn{3}{|c||}{Poker Hand} & \multicolumn{3}{|c}{Gas Turbine CO and NOx Emission} \\
				\cline{2-10}
				& $sr = 0.2\%$ & $sr = 0.5\%$ & $sr = 1\%$ & $sr = 0.1\%$ & $sr = 0.5\%$ & $sr = 1\%$ & $sr = 0.1\%$ & $sr = 0.5\%$ & $sr = 1\%$ \\
				\hline
				UNIF & 37.375(757.056) & 17.544(150.132) & 14.091(10.515) & 20.628(39.454) & 16.494(3.870) & 16.145(1.733) & 1.515(2.641) & 1.212(0.268) & 1.187(0.116) \\
				LEV & 977.049(1113.369) & 30.047(832.387) & 14.793(34.945) & 21.687(86.551) & 16.464(4.402) & 16.121(1.558) & 1.401(1.556) & 1.198(0.147) & 1.182(0.074) \\
				SLEV1 & 43.437(1145.047) & 32.101(786.459) & 17.922(145.850) & 20.199(33.124) & 16.462(3.625) & 16.143(1.602) & 1.477(2.216) & 1.204(0.170) & 1.187(0.127) \\
				SLEV5 & 716.760(1126.215) & 39.196(1336.134) & 17.582(199.610) & 21.613(52.904) & 16.501(3.955) & 16.118(1.350) & 1.428(2.594) & 1.198(0.155) & 1.181(0.072) \\
				SLEV9 & 759.980(1476.744) & 19.909(286.266) & 18.714(242.235) & 20.688(41.378) & 16.435(2.933) & 16.163(1.864) & 1.386(1.148) & 1.198(0.129) & 1.182(0.095) \\
				LEVUNW & 23.865(118.949) & 15.038(9.115) & 13.849(3.739) & 20.002(33.140) & 16.448(3.846) & 16.172(1.984) & \textbf{1.355}(0.935) & 1.209(0.182) & 1.192(0.083) \\
				GS & 959.747(1093.527) & 20.077(239.837) & 21.219(754.344) & 22.568(66.602) & 16.425(3.497) & 16.057(1.138) & 1.523(3.019) & \textbf{1.195}(0.145) & 1.180(0.082) \\
				IS & 773.265(1675.704) & 79.019(3602.502) & 57.206(2274.425) & 22.089(55.309) & 16.391(2.900) & 16.044(0.968) & 1.513(2.485) & 1.203(0.175) & 1.180(0.069) \\
				HMS & \textbf{21.906}(106.737) & \textbf{14.267}(15.968) & \textbf{13.549}(6.247) & \textbf{18.594}(15.623) & \textbf{16.217}(1.871) & \textbf{16.022}(1.044) & 1.365(1.164) & 1.197(0.161) & \textbf{1.178}(0.052) \\
				\hline
			\end{tabular}
		}
	\end{center}
	\caption{APE comparisons (mean $\pm$ standard deviation) for different sampling methods for real datasets}
	\label{realdata1}
\end{table*}

We compare the proposed HMS with several representative methods, including uniform sampling (UNIF), leverage subsampling (LEV) \cite{drineas2012fast}, unweighted leverage subsampling (LEVUNW), shrinkage leverage subsampling (SLEV) \cite{ma2015statistical},  gradient-based sampling (GS) \cite{zhu2016grad} and influence-based sampling (IS) \cite{ting2018optimal}. The sampling probability of SLEV is a convex combination of leverage and uniform distribution, i.e. 
$ \pi_i^{SLEV} = \alpha\pi_i^{LEV} + (1 - \alpha)\pi_i^{UNIF}$. Here, we consider 3 different shrinkage factors $\alpha = 0.1, 0.5, 0.9$ for SLEV, denoted by SLEV0.1, SLEV0.5 and SLEV0.9 respectively.  LEVUNW performs the same sampling procedure as LEV, but solves the unweighted least squares problem
instead. For influence-based sampling, the sampling weight for $(\x_i. y_i)$ is proportional to $\|\psi_{\bbeta} (\x_i, y_i)\|$, where
$\psi_{\bbeta} (\x_i, y_i) = (y_i - \x_i \bbeta) \sum_n^{-1} \x_i$
is the influence function. For GS, IS and HMS, the pilot is calculated by uniform sampling with size $n_0 = n_{sub}$, the parameter $\tau$ in HMS is specified through a grid search strategy. 

For each model, we set $n =100K, 1M$ and corresponding $d =50, 500$. Denote $sr$ by the sampling ratio, we set subsample size by $n_{sub} = sr * n$ with $sr = 0.001, 0.005, 0.01$.  Each result is reported over $K = 100$ runs repeatedly and the mean error is calculated.

The AME comparisons for different sampling procedures are demonstrated in  Figures \ref{fig_line} - \ref{fig_boxplot4}, and the corresponding running time comparison is shown in Table \ref{tab_time}. Several observations can be made about the reported results. (I) Leverage based sampling procedures perform slightly worse than uniform sampling when data are corrupted by heavy tailed noises, this is because leverage can not exactly reflect the true importance of each sample in such cases. (II) GS and IS behave similar in different settings. The reason is that the design matrix $\X$ consists of a mixture of i.i.d. Gaussian entries,  leading to the covariance matrix $\Sigma_n$ approximates a diagonal matrix, which makes influence function assigns similar sampling probability as gradient does. (III) GS and IS perform worse than leverage based approaches and uniform sampling when sampling ratio is small. The main reason is that both of them need a pilot to guide sampling, inefficient training for the pilot will deteriorate their performance. However, HMS performs significantly better than GS and IS with the same pilot. This demonstrates the tolerance of HMS to imperfect pilots. (IV) HMS performs much better than the other competitors in almost all settings, both in AME and running time. The efficiency improvement of HMS is still prominent even considering the time for hyper-parameter ($\tau$) selection, which implies the great advantage of HMS on selecting the informative samples under high level noise settings.

\begin{table*}[h]
	
	\begin{center}
		\resizebox{\textwidth}{14mm}{
			\footnotesize
			\begin{tabular}{ c||ccc||ccc||ccc }
				\hline
				\multirow{2}{4em}{Methods} & \multicolumn{3}{|c||}{Wave Energy Converters} & \multicolumn{3}{|c||}{Physicochemical Properties of Protein Tertiary Structure} & \multicolumn{3}{|c}{Beijing Multi-Site Air-Quality Data} \\
				\cline{2-10}
				& $sr = 0.1\%$ & $sr = 0.5\%$ & $sr = 1\%$ & $sr = 0.1\%$ & $sr = 0.5\%$ & $sr = 1\%$ & $sr = 0.1\%$ & $sr = 0.5\%$ & $sr = 1\%$ \\
				\hline
				UNIF & 55.611(9.120) & 52.892(1.638) & 52.594(0.680) & 22.946(33.494) & 19.074(8.027) & 18.647(4.511) & 27.930(9.177) & 27.207(1.000) & 27.125(0.330) \\
				LEV & 55.553(9.456) & 52.875(1.724) & 52.585(0.783) & 20.328(11.959) & 18.494(2.109) & 18.298(0.891) & 27.535(2.552) & 27.162(0.455) & 27.109(0.205) \\
				SLEV1 & 55.305(8.762) & 52.866(1.634) & 52.585(0.665) & 22.336(49.774) & 18.509(3.508) & 18.299(0.915) & 27.698(4.926) & 27.176(0.541) & 27.118(0.350) \\
				SLEV5 & 55.528(8.142) & 52.853(1.439) & 52.591(0.826) & 20.207(13.829) & 18.399(1.568) & 18.269(0.842) & 27.586(2.942) & 27.156(0.467) & 27.108(0.301) \\
				SLEV9 & 55.425(9.317) & 52.885(1.679) & 52.579(0.795) & 20.024(12.033) & 18.446(1.916) & 18.295(0.834) & 27.523(2.037) & 27.158(0.457) & 27.106(0.208) \\
				LEVUNW & 55.365(6.857) & 52.906(1.462) & 52.625(0.786) & 19.562(6.400) & 18.700(1.997) & 18.558(1.131) & 27.832(6.767) & 27.338(1.295) & 27.259(0.736) \\
				GS & 55.905(10.413) & \textbf{52.753}(1.170) & \textbf{52.504}(0.528) & 21.165(16.045) & 18.463(1.907) & 18.265(0.872) & 27.346(1.187) & 27.100(0.157) & \textbf{27.076}(0.066) \\
				IS & 55.738(9.943) & 52.756(1.259) & 52.517(0.581) & 21.364(20.183) & 18.474(1.561) & 18.427(2.645) & 27.343(1.236) & 27.104(0.172) & 27.078(0.082) \\
				HMS & \textbf{54.982}(6.715) & 52.789(1.323) & 52.560(0.683) & \textbf{19.472}(15.802) & \textbf{18.299}(1.580) & \textbf{18.089}(0.871) & \textbf{27.185}(0.479) & \textbf{27.085}(0.075) & 27.089(0.063) \\
				\hline
			\end{tabular}
		}
	\end{center}
	\caption{APE comparisons (mean $\pm$ standard deviation) for different sampling methods for real datasets}
	\label{realdata2}
\end{table*}

\subsection{Real Data Examples}

We further evaluate the proposed HMS on 6 real-world datasets. Including \textit{Appliances Energy Prediction}, \textit{Poker Hand}, \textit{Gas Turbine CO and NOx Emission }, \textit{Wave Energy Converters}, \textit{Physicochemical Properties of Protein Tertiary Structure (PPPTS) } and \textit{Beijing Multi-Site Air-Quality}. All these datasets come from UCI machine learning repository \url{https://archive.ics.uci.edu/ml/datasets.php}, covering various prediction tasks. For \textit{Poker Hand} dataset, we only use the training set. For \textit{Wave Energy Converters} dataset, we remove 16 columns due to collinearity. For \textit{Beijing Multi-Site Air-Quality} dataset, we remove 4 text-valued columns, and take \textit{PM2.5} as the prediction target.  The results are averaged over $K = 100$ runs of each experiment, and the average prediction errors (APE):
\begin{equation*}
	\textrm{APE} = \frac{1}{K} \sum_{k=1}^{K} \| \hat{\y}_k - \y\| 
\end{equation*}
are reported in Table \ref{realdata1} and \ref{realdata2}. It can be observed that  HMS can achieve superior performance in these regression tasks. Specifically, HMS almost always reach the lowest error and standard deviation when sampling ratio remains small, this shows the great potential of applying HMS to deal with big data. For the Gas Emission, Wave Energy and Air-Quality datasets, HMS sometimes yields sub-optimal results comparing to other methods. This is because that in real-world scenarios, the properties of the noise are unknown, and some of the assumptions are not guaranteed to be hold, i.e. bounded covariates or bounded $1 + \delta$ order error moments. The convergence of HMS is thus influenced and results in sub-optimal samples. However, HMS still achieves the highest performance in most conditions, which demonstrates its outstanding robustness over other methods.

\section{Discussion and Future Research}\label{sec:con}
In this paper, we propose a Markov subsampling strategy based on Huber criterion (HMS) to achieve robust estimation. The deviation bounds of HMS estimator are established. We find that the HMS estimator exhibits a similar phase transition to that in the independent setup. The only difference is up to a factor $\sqrt{\frac{1 -\lambda}{1 + \lambda}}$, defined by the absolute spectral gap  $\lambda$ of underlying Markov chain.  
Extensive studies on large scale simulations and  real data examples demonstrate the effectiveness of HMS. There are many opportunities along the line of current research, such as   how to deduce the lower bounds for HMS estimator and how to perform HMS in high dimensional case. All these problems deserve further research.

%\section*{Acknowledgment}

\bibliographystyle{IEEEtran}
\bibliography{IEEEfull,hms}

\newpage
\appendix

\section{Supplementary Experimental Results}
\begin{figure*}[ht]
	{\footnotesize \ \centering
		\subfigure{
			\includegraphics[scale=0.35]{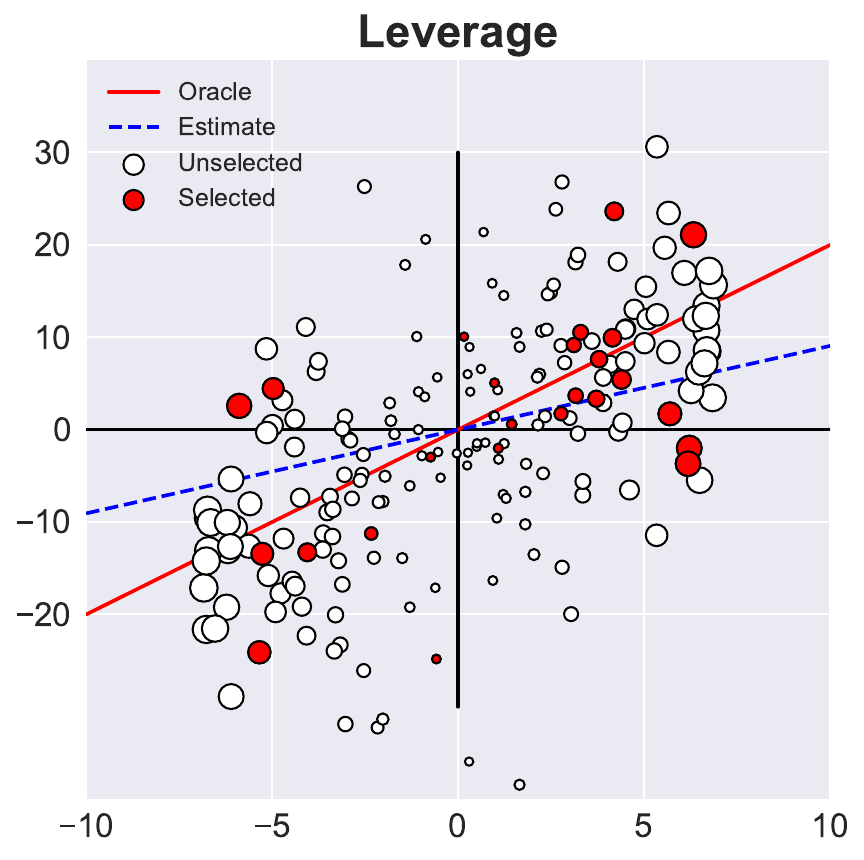}}
		\subfigure{
			\includegraphics[scale=0.35]{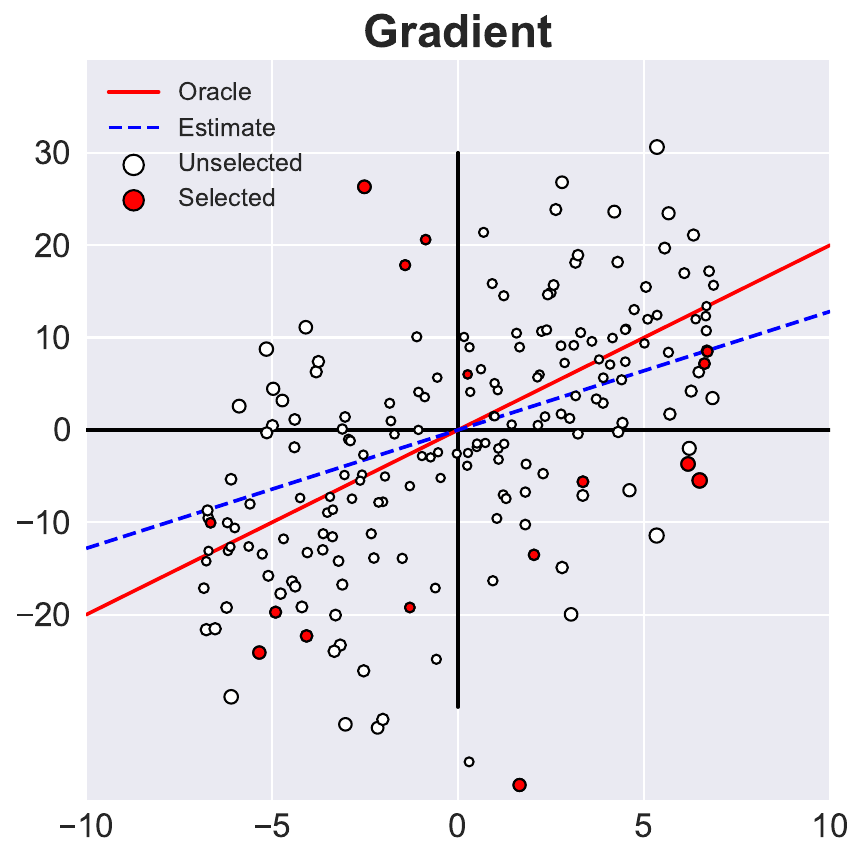}}\\
		\subfigure{
			\includegraphics[scale=0.35]{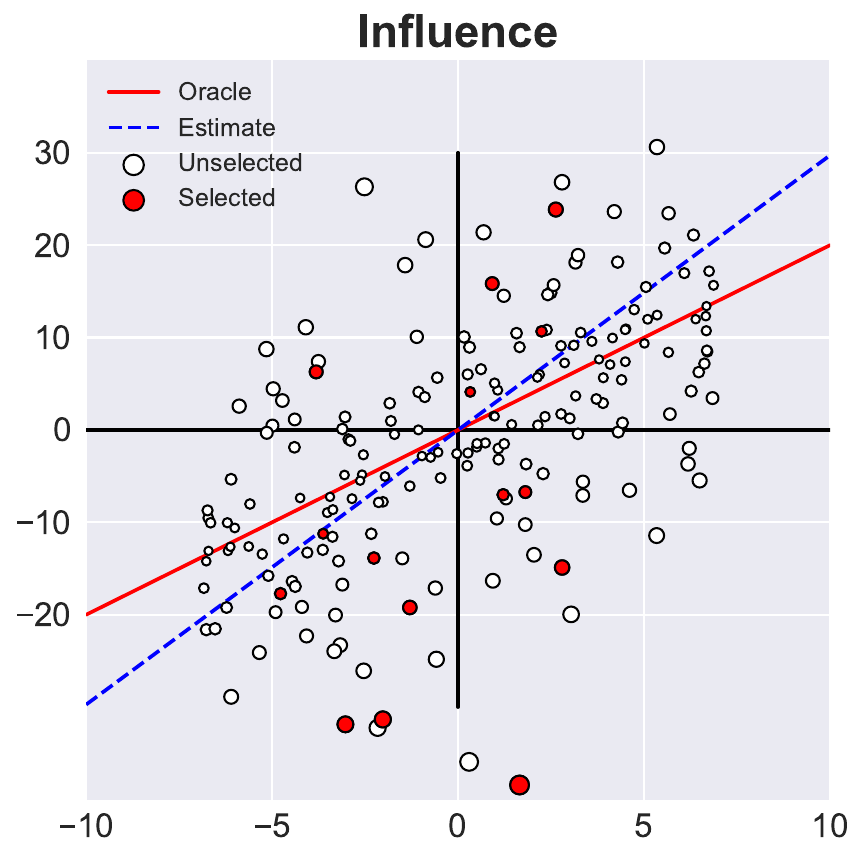}}
		\subfigure{
			\includegraphics[scale=0.35]{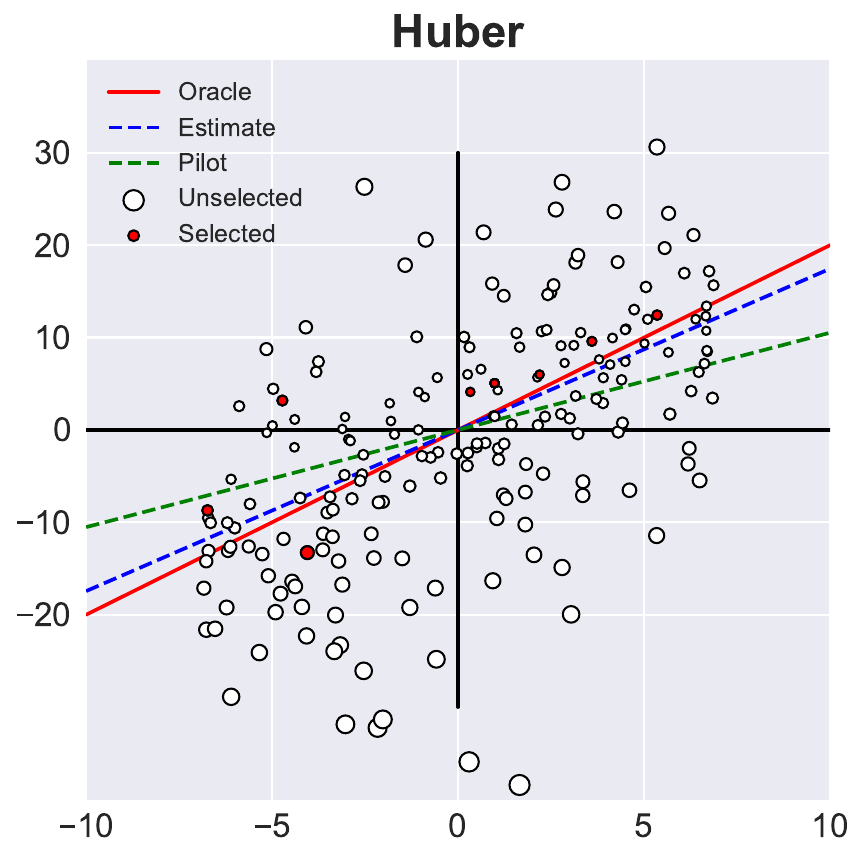}} 
		\caption{ Comparisons on different sampling patterns with $n = 200$ and $\varepsilon \sim \mathbf{t}(2)$. }
		\label{fig_patt1}
	}
\end{figure*}

\begin{figure*}[ht]
	{\footnotesize \ \centering
		\subfigure{
			\includegraphics[scale=0.35]{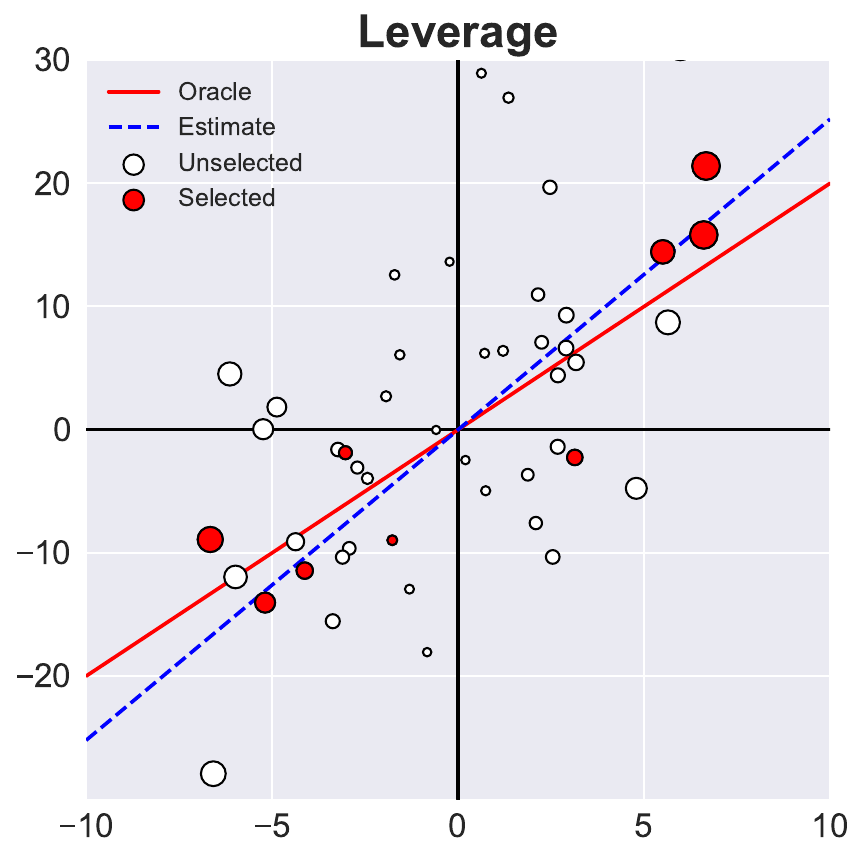}}
		\subfigure{
			\includegraphics[scale=0.35]{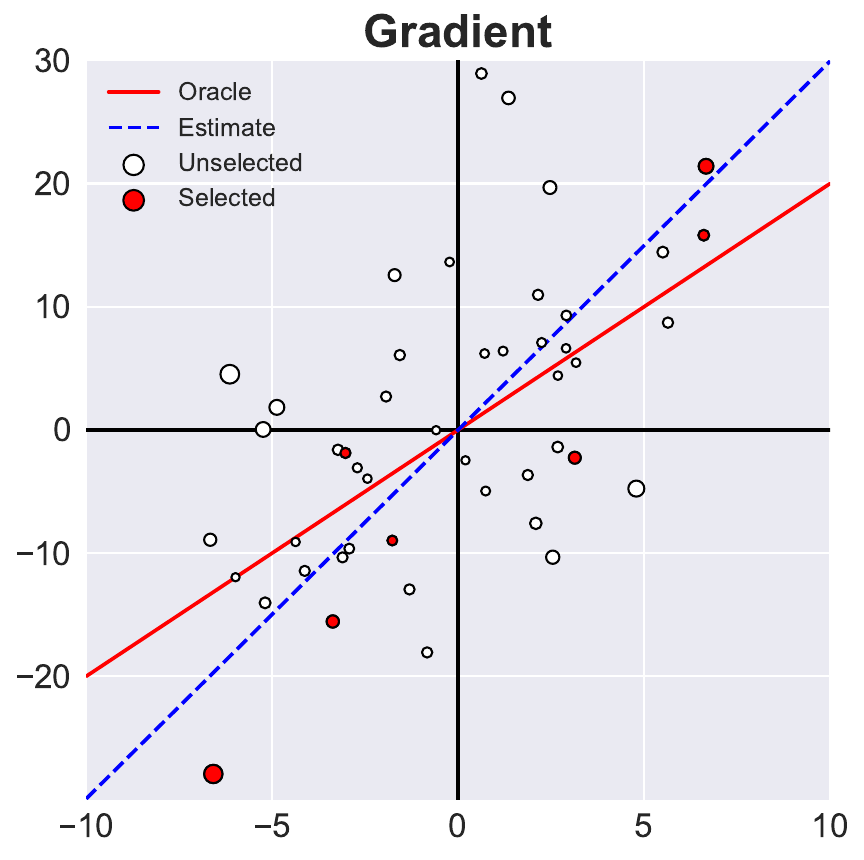}}\\
		\subfigure{
			\includegraphics[scale=0.35]{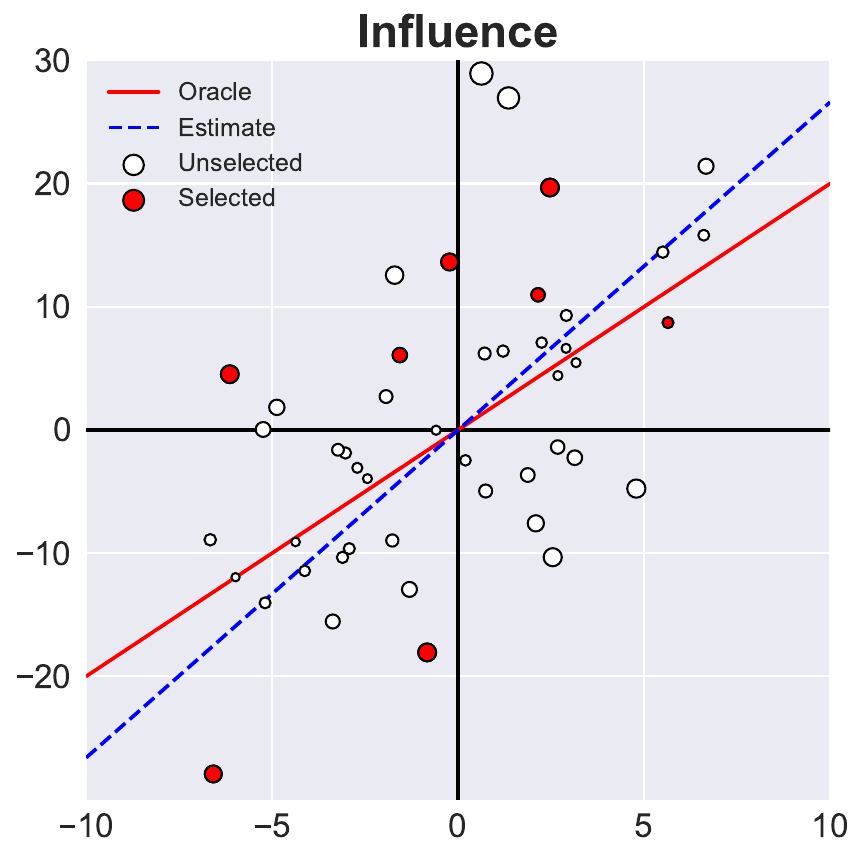}}
		\subfigure{
			\includegraphics[scale=0.35]{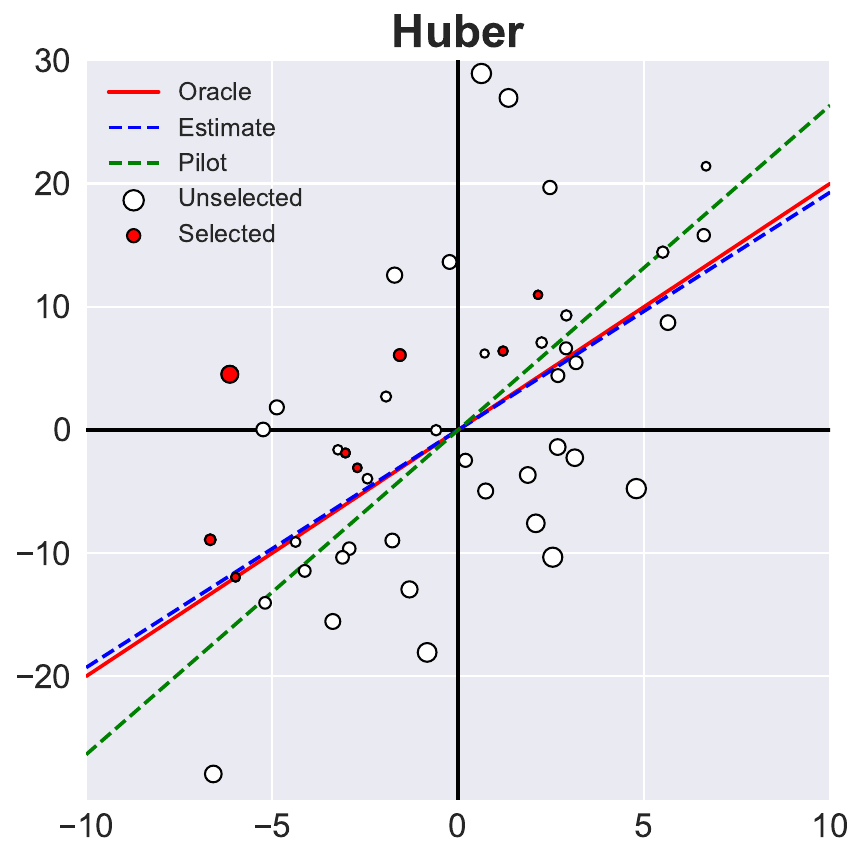}} 
		\caption{ Comparisons on different sampling patterns with $n = 50$ and $\varepsilon \sim \mathbf{Lognormal}(0, 1)$. }
		\label{fig_patt2}
	}
\end{figure*}

\begin{figure*}[ht]
	{\footnotesize \ \centering
		\subfigure{
			\includegraphics[scale=0.35]{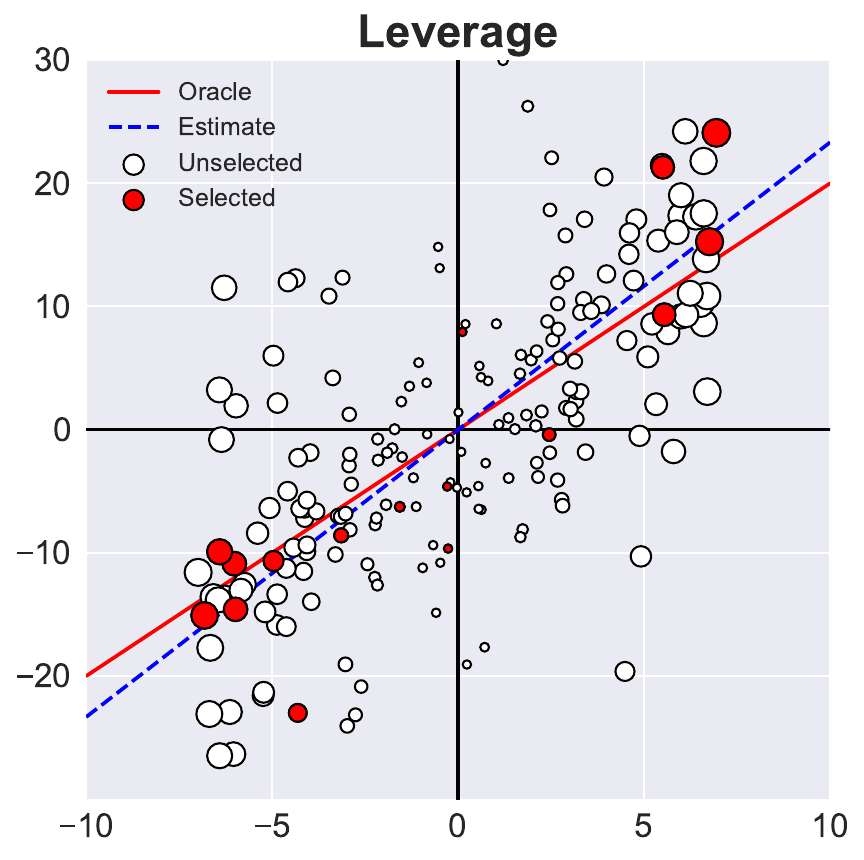}}
		\subfigure{
			\includegraphics[scale=0.35]{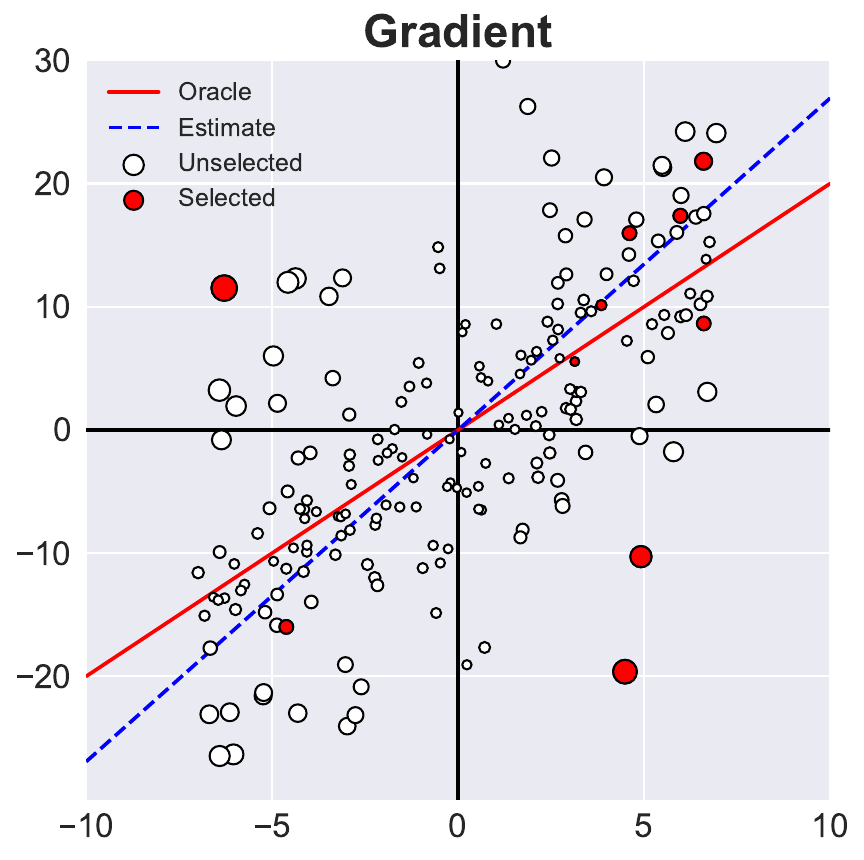}}\\
		\subfigure{
			\includegraphics[scale=0.35]{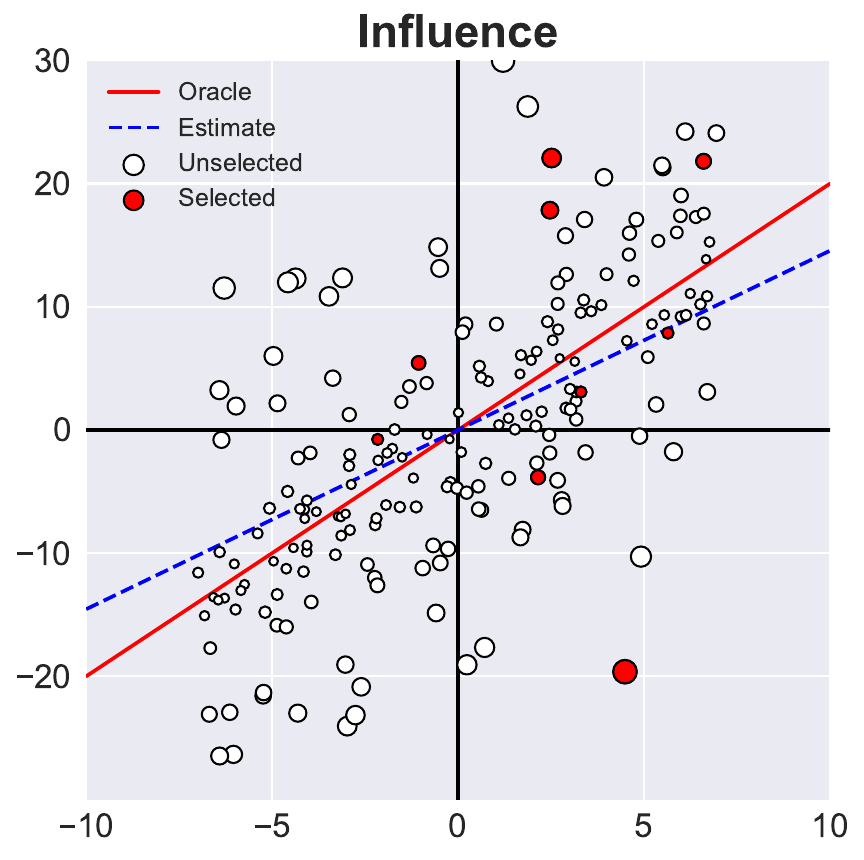}}
		\subfigure{
			\includegraphics[scale=0.35]{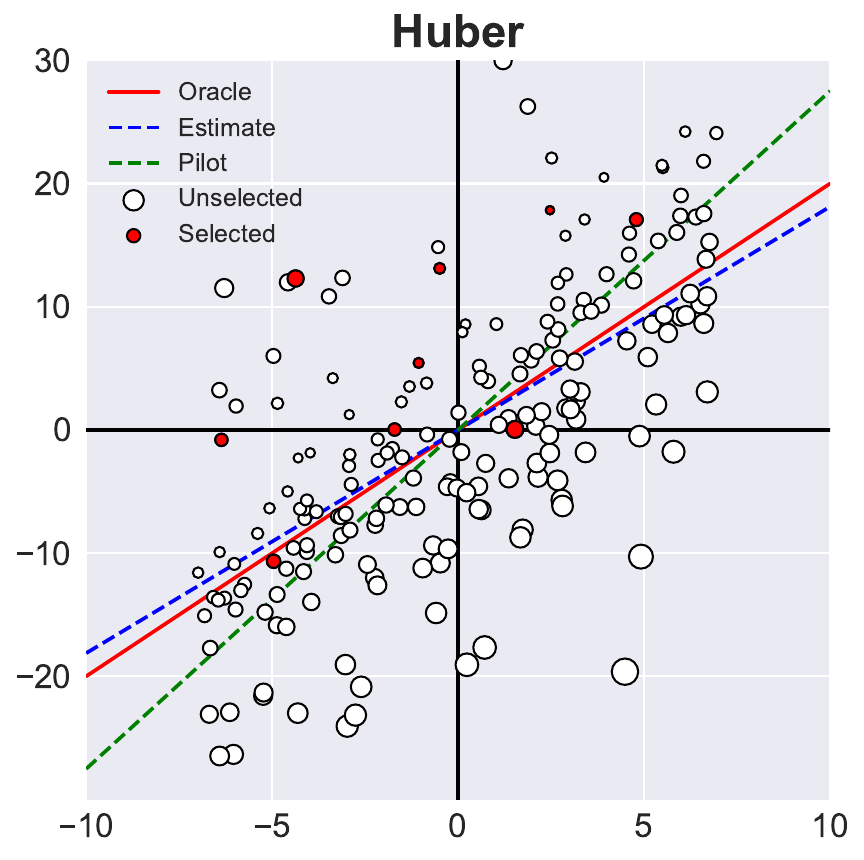}} 
		\caption{ Comparisons on different sampling patterns with $n = 200$ and $\varepsilon \sim \mathbf{Lognormal}(0, 1)$. }
		\label{fig_patt3}
	}
\end{figure*}

In this section, we add supplementary experiments on different data scales. In Figure \ref{fig_patt1}, \ref{fig_patt2} and \ref{fig_patt3}, we give additional experiment results of sampling patterns with different subsampling strategies. We keep the same settings as ``A. Sampling Pattern'' and set number of samples $n = \{50, 200\}$, distribution of noises $\varepsilon \sim \{\mathbf{t}(2), \mathbf{Lognormal}(0, 1)\}$. In all of these settings, we can derive the same conclusion that HMS achieves the lowest estimation error, even the pilot is deviated from the oracle.

\begin{figure*}[ht]
	\centering
	\includegraphics[scale=0.5]{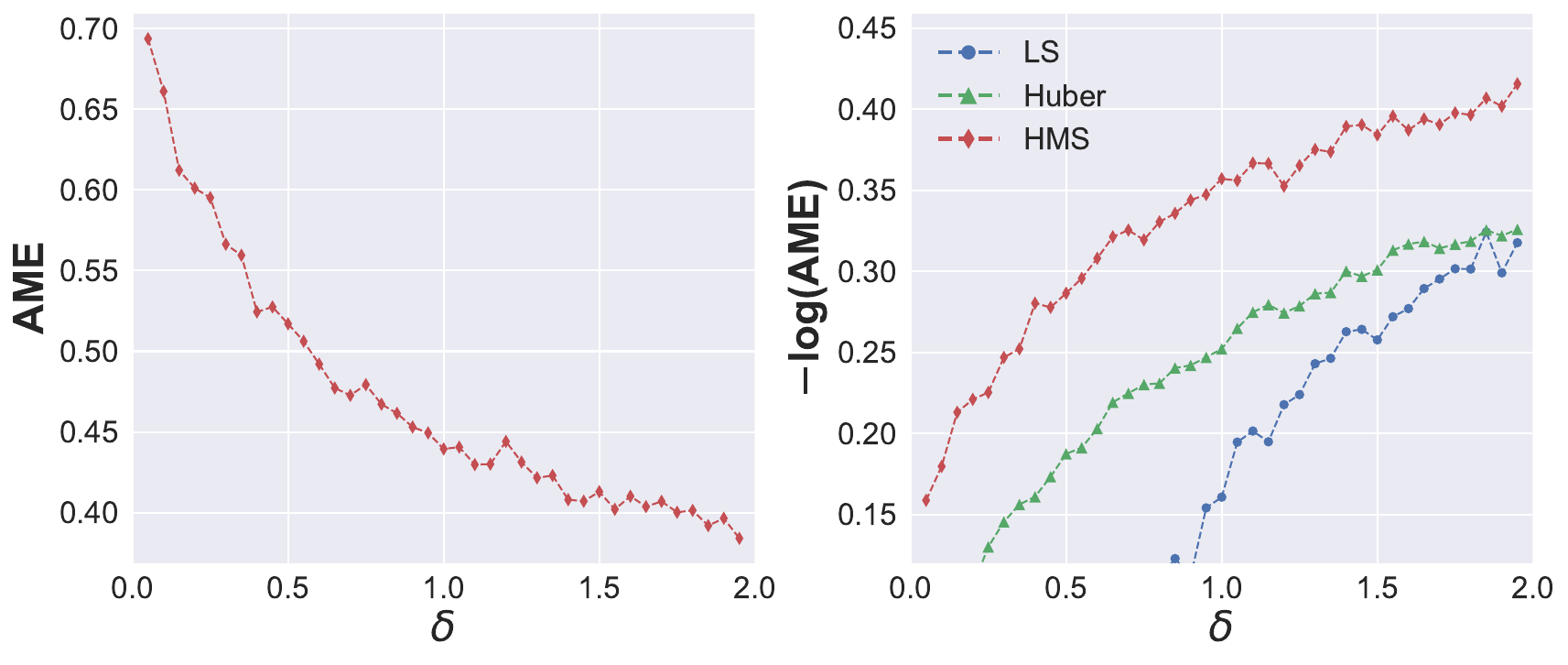}
	\caption{Comparisons on AME of different sampling procedures with $n = 5000$, $d = 25$ and $n_{sub} = 50$.}
	\label{fig_phase1}
\end{figure*}

\begin{figure*}[ht]
	\centering
	\includegraphics[scale=0.5]{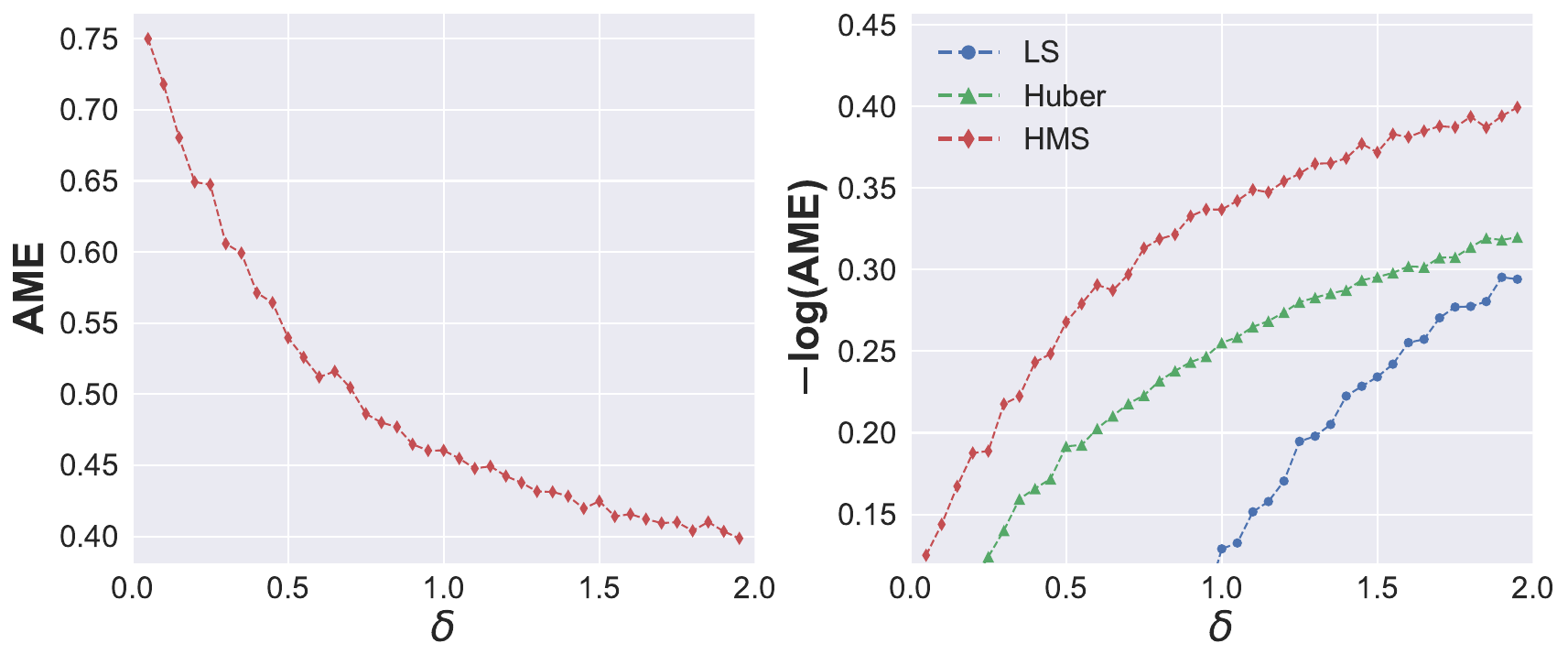}
	\caption{Comparisons on AME of different sampling procedures $n = 20000$, $d = 100$ and $n_{sub} = 200$.}
	\label{fig_phase2}
\end{figure*}

In Figure \ref{fig_phase1} and \ref{fig_phase2}, we give additional experiment results of the phase transition behavior. Again, we keep the same parameter settings as ``B. Phase Transition'' and alter the simulation data size to $n = 5000, d = 25, n_{sub} = 50$ and $n = 20000, d = 100, n_{sub} = 200$. As can be seen, HMS still achieves lower AME than Huber and LS consistently under various data scales.

\end{document}